\documentclass{article}

\PassOptionsToPackage{numbers, compress}{natbib}
\usepackage[final]{neurips_2020}
\usepackage[utf8]{inputenc} % allow utf-8 input
\usepackage[T1]{fontenc}    % use 8-bit T1 fonts
\usepackage{hyperref}       % hyperlinks
\usepackage{url}            % simple URL typesetting
\usepackage{booktabs}       % professional-quality tables
\usepackage{amsfonts}       % blackboard math symbols
\usepackage{nicefrac}       % compact symbols for 1/2, etc.
\usepackage{microtype}      % microtypography
\usepackage{multirow}
\usepackage{array}
\usepackage{graphicx}
\usepackage{color}
\usepackage{xcolor}
\usepackage{paralist}
\usepackage{bbm}
\usepackage{amsmath}
\usepackage{algorithm}
\usepackage{algpseudocode}
\usepackage{graphics}
\usepackage{epsfig}
\usepackage{wrapfig}
\usepackage{caption}
\usepackage{makecell}
\usepackage{stackengine}
\usepackage[marginal]{footmisc}

\newcommand{\mypar}[1]{{\bf #1.}}
\newcommand{\tabincell}[2]{\begin{tabular}{@{}#1@{}}#2\end{tabular}}

\title{Graph Cross Networks with Vertex Infomax Pooling}

\author{
Maosen Li \\
Shanghai Jiao Tong University \\
\texttt{maosen\_li@sjtu.edu.cn}
\And
Siheng Chen \\
Mitsubishi Electric Laboratories \\
\texttt{schen@merl.com}
\And
Ya Zhang \\
Shanghai Jiao Tong University \\
\texttt{ya\_zhang@sjtu.edu.cn}
\And 
Ivor Tsang \\
University of Technology, Sydney \\
\texttt{Ivor.Tsang@uts.edu.au}
}

\begin{document}

\maketitle
% \tableofcontents

\begin{abstract}
We propose a novel~\emph{graph cross network} (GXN) to achieve comprehensive feature learning from multiple scales of a graph. Based on trainable hierarchical representations of a graph, GXN enables the interchange of intermediate features across scales to promote information flow. Two key ingredients of GXN include a novel~\emph{vertex infomax pooling} (VIPool), which creates multiscale graphs in a trainable manner, and a novel feature-crossing layer, enabling feature interchange across scales.  The proposed VIPool selects the most informative subset of vertices based on the neural estimation of mutual information between vertex features and neighborhood features. The intuition behind is that a vertex is informative when it can maximally reflect its neighboring information. The proposed feature-crossing layer fuses intermediate features between two scales for mutual enhancement by improving information flow and enriching multiscale features at hidden layers. The cross shape of feature-crossing layer distinguishes GXN from many other multiscale architectures. Experimental results show that the proposed GXN improves the classification accuracy  by $2.12\%$ and $1.15\%$ on average for graph classification and vertex classification, respectively. Based on the same network, the proposed VIPool consistently outperforms other graph-pooling methods.

\end{abstract}
\vspace{-10pt}

\section{Introduction}
\vspace{-1mm}
Recently, there are explosive interests in studying graph neural networks (GNNs)~\cite{ICLR2017_Kipf,NIPS2017_6703,velickovic2018graph,NIPS2016_Kipf,ICML_2016_Dai,NIPS2015_5954,AAAI1817146,NIPS2018_7729,ICML2019_Lee,Li_2019_CVPR}, which expand deep learning techniques to ubiquitous non-Euclidean graph data, such as social networks~\cite{NIPS2015_5880}, bioinformatic networks~\cite{DD} and human activities~\cite{Li_2019_CVPR}. 
Achieving good performances on graph-related tasks, such as vertex classification~\cite{ICLR2017_Kipf,NIPS2017_6703,velickovic2018graph} and graph classification~\cite{NIPS2015_5954,AAAI1817146,NIPS2018_7729}, GNNs learn patterns from both graph structures and vertex information with feature extraction in spectral domain~\cite{Bruna2014ICLR,NIPS2016_6081,ICLR2017_Kipf} or vertex domain~\cite{NIPS2017_6703,pmlr-v70-Niepert16,velickovic2018graph,pmlr-v70-lei17a,Li2016ICLR,ICML_2016_Dai,963781}. Nevertheless, most GNN-based methods learn features of graphs with fixed scales, which might underestimate either local or global information. To address this issue, multiscale feature learning on graphs enables capturing more comprehensive graph features for downstream tasks~\cite{AAAI1816273,ICML2019_Gao,liao2018lanczosnet}. 

{Multiscale feature learning on graphs is a natural generalization from multiresolution analysis of images, whose related techniques, such as wavelets and pyramid representations, have been well studied in both theory and practice~\cite{He_2019_ICCV,Sun_2019_CVPR,Ronneberger_2015_MICCAI,7803544,Zhang_2017_ICCV}.} 
% Compared to the grid-like images, graphs are usually highly irregular, causing two main issues to model the multiscale graph for feature learning: 
{However, this generalization is technically nontrivial. While hierarchical representations and pixel-to-pixel associations across scales are straightforward for images with regular lattices, the highly irregular structures of graphs cause challenges in producing graphs at various scales~\cite{ChenVSK:15} and aggregating features across scales.}

To generate multiscale graphs, graph pooling methods are essential to compress large graphs into smaller ones. Conventional graph pooling methods~\cite{ChenVSK:15,SafroS:14} leverage graph sampling theory and designed rules. Recently, some data-driven pooling methods are proposed, which automatically merge a fine-scale vertex subset to a coarsened vertex~\cite{4302760,NIPS2016_6081,Simonovsky_2017_CVPR,ijcai2018-490,DBLP:journals/corr/abs-1802-09612,NIPS2018_7729}. The coarsened graph, however, might not have direct vertex-to-vertex association with the original scale. Some other graph pooling methods adaptively select vertices based on their importance over the entire graph~\cite{ICML2019_Gao,ICML2019_Lee}; however, they fail to consider local information.

To aggregate features across multiple scales, existing attempts build encoder-decoder architecture~\cite{AAAI1816273,DBLP:journals/corr/abs-1802-09612,ICLR2020_Deng} to learn graph features from the latent spaces, which might underestimate fine-scale information. Some other works gather the multiscale features in parallel and merge them as the final representation~\cite{liao2018lanczosnet,ICML2019_Gao,DBLP:journals/corr/abs-1902-06684,ICML2019_Lee}, which might limit information flow across scales.

In this work, we design a new graph neural network to achieve multiscale feature learning on graphs, and our technical contributions are two-folds: a novel graph pooling operation to preserve informative vertices and a novel model architecture to exploit rich multiscale information.

\begin{figure*}[!t]
  \vspace{-5pt}
  \begin{center}
    \begin{tabular}{cccc}
     \includegraphics[width=0.22\columnwidth]{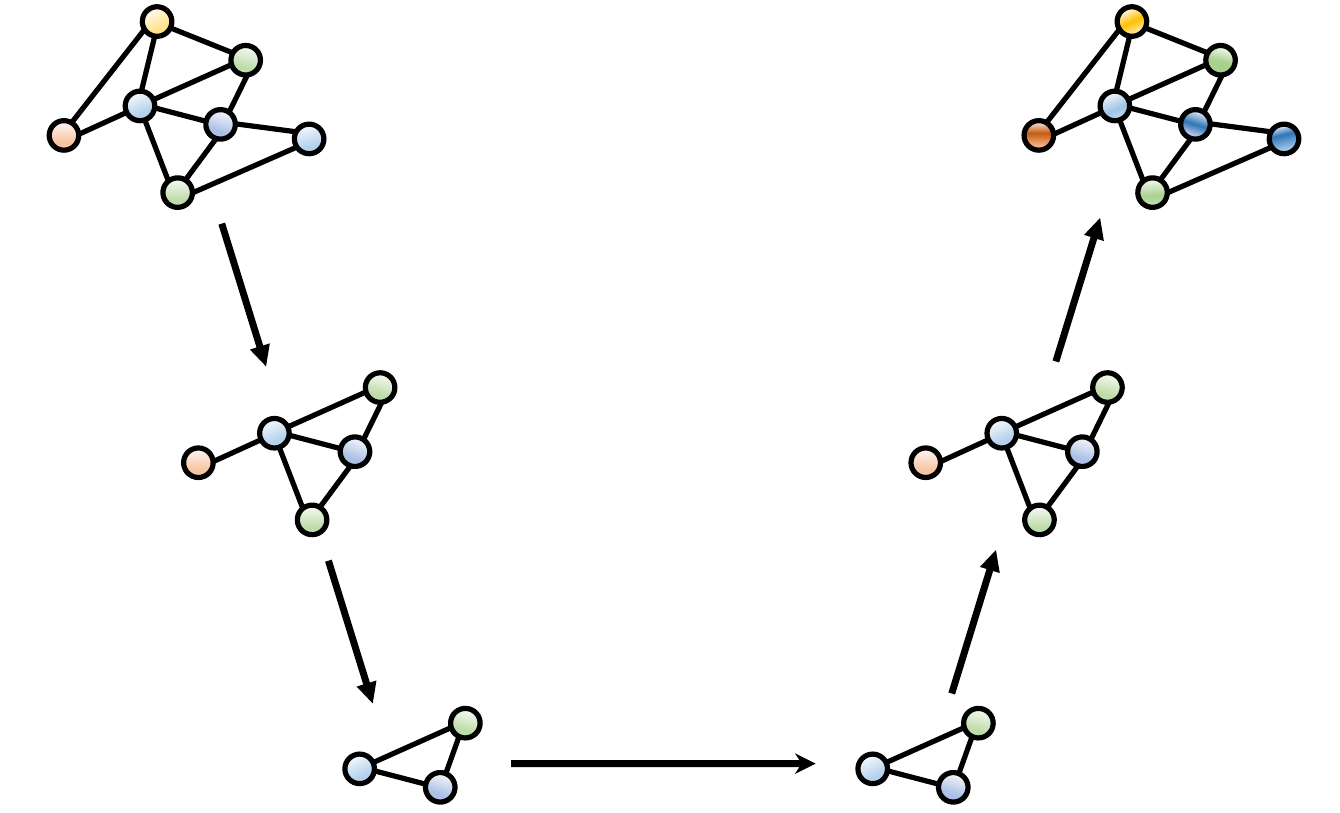}   
   & \includegraphics[width=0.22\columnwidth]{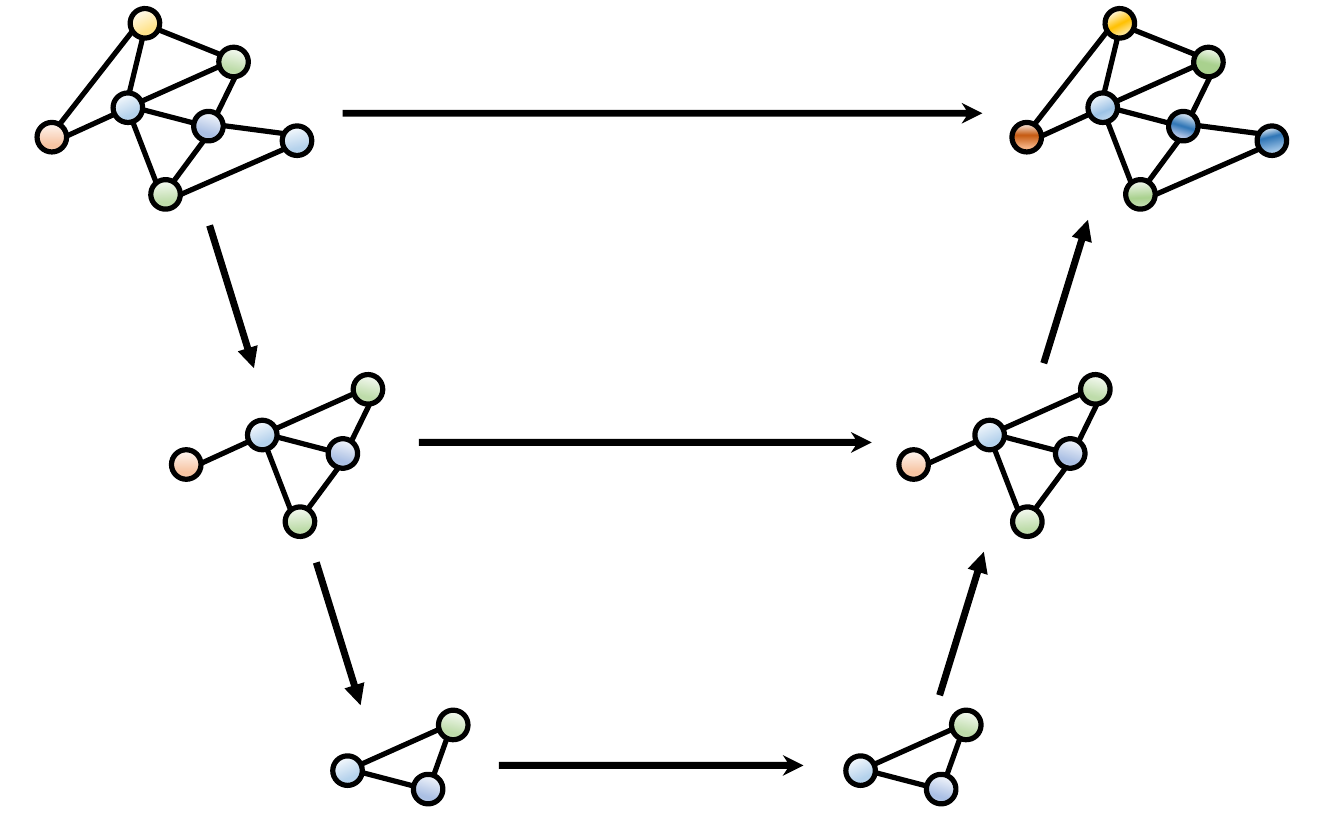} 
   & \includegraphics[width=0.217\columnwidth]{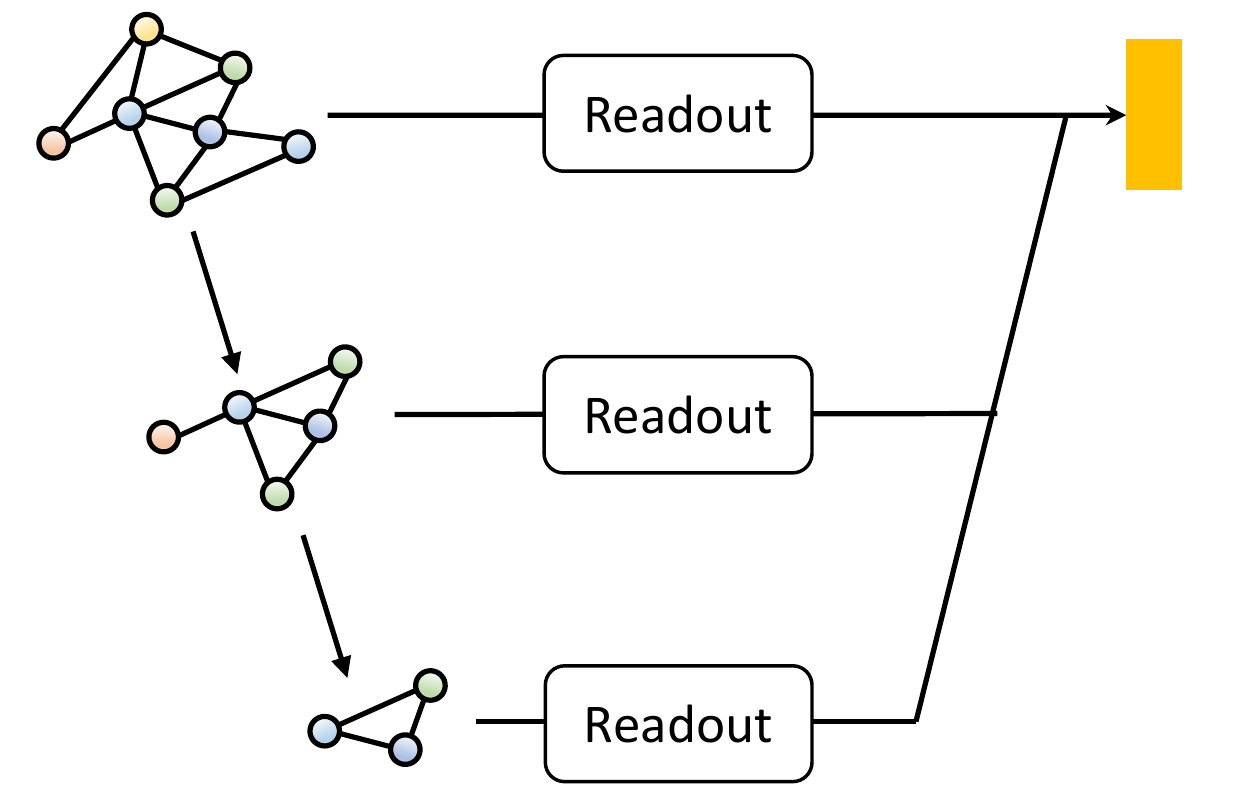}
   & \includegraphics[width=0.22\columnwidth]{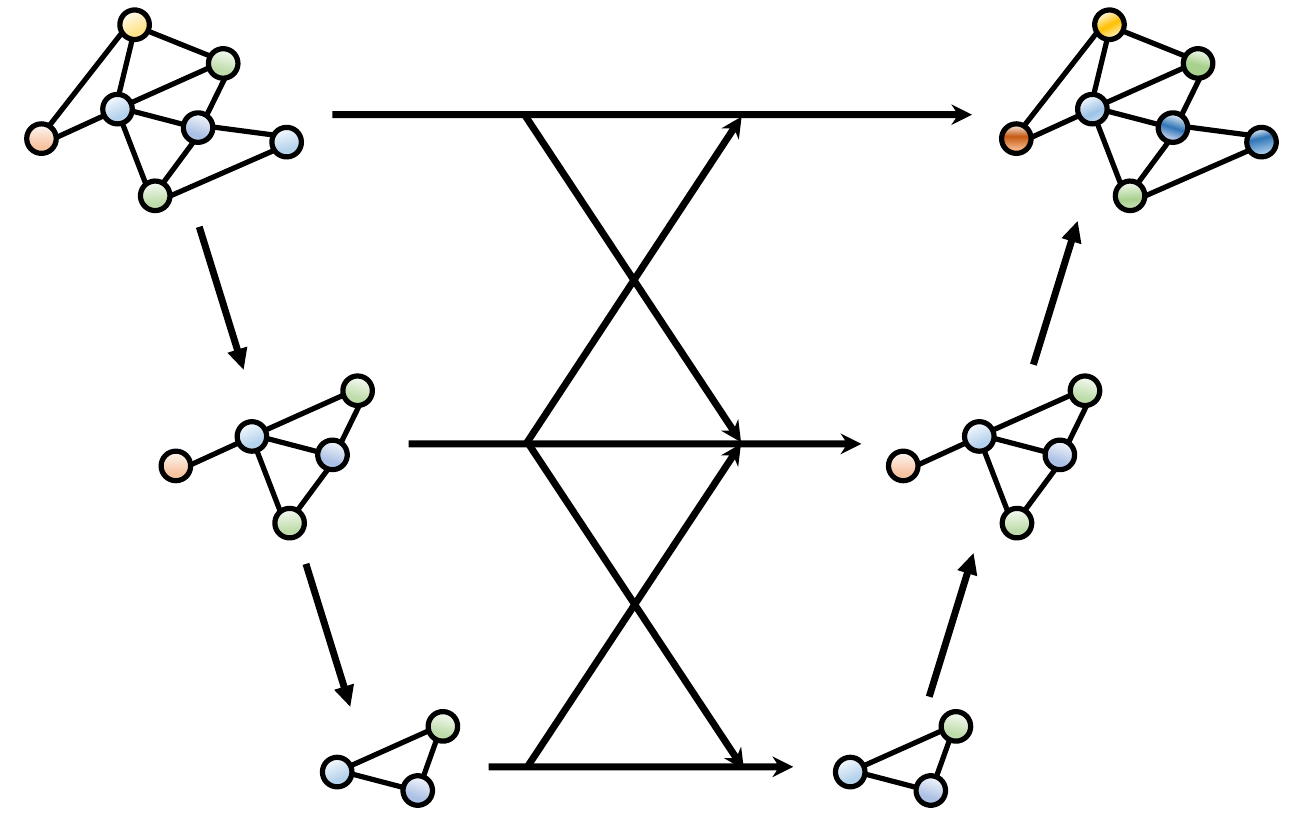}
       \\
    {\small (a) Encoder-decoder~\cite{AAAI1816273}.} &  
    {\small (b) Graph U-net~\cite{ICML2019_Gao}.} &
    {\small (c) Readout~\cite{ICML2019_Lee}.} &
    {\small (d) GXN (ours).}
  \end{tabular}
\end{center}
\vspace{-10pt}
\caption{\small Architectures of multiscale graph neural networks. Our architecture adopts intermediate fusion.}
\vspace{-15pt}
\label{fig:architecture_comparison}
\end{figure*}

\mypar{A novel graph pooling operation:  Vertex infomax pooling (VIPool)} 
% We propose a novel graph pooling operation called~\emph{vertex infomax pooling} (VIPool), which select and preserve those vertices that can maximally express their corresponding neighborhoods based neural estimation of mutual information~\cite{pmlr-v80-belghazi18a,hjelm2019learning,velickovic2019deep} between vertices and their neighborhood.
We propose a novel graph pooling operation by selecting and preserving those vertices that can maximally express their corresponding neighborhoods. The criterion of vertex-selection is based on the neural estimation of mutual information~\cite{pmlr-v80-belghazi18a,hjelm2019learning,velickovic2019deep} between vertex and neighborhood features, thus we call the proposed pooling mechanism~\emph{vertex infomax pooling} (VIPool). 
Based on VIPool, we can implement graph pooling and unpooling to coarsen and refine multiple scales of a graph. Compared to the vertex-grouping-based methods~\cite{4302760,NIPS2016_6081,Simonovsky_2017_CVPR,ijcai2018-490,DBLP:journals/corr/abs-1802-09612,NIPS2018_7729}, the proposed VIPool provides the direct vertex-vertex association across scales and makes the coarsened graph structure and information fusion easier to achieve. Compared to other vertex-selection-based methods~\cite{ICML2019_Gao,ICML2019_Lee}, VIPool considers both local and global information on graphs by learning both vertex representation and graph structures.

\mypar{A novel model architecture: Graph cross network (GXN)} 
% The proposed GXN model is constructed with a novel architecture that captures rich multiscale information for graph representation learining.
We propose a new model with a novel architecture called \emph{graph cross network} (GXN) to achieve feature learning on multiscale graphs. 
Employing the trainable VIPool, our model creates multiscale graphs in data-driven manners. To learn features from all parallel scales, our model is built with a pyramid structure. To further promote information flow, we propose novel intermediate \emph{feature-crossing layers} to interchange features across scales in each network layer. The intuition of feature-crossing is that the it improves information flow and exploits richer multiscale information in multiple network layers rather than only combine them in the last layer.  Similar crossing 
structures have been explored for analyzing images~\cite{7849978,Sun_2019_CVPR}, but we cannot directly use those structures for irregular graphs. The proposed feature-crossing layer handles irregular graphs by providing the direct vertex-vertex associations across multiple graph scales and network layers; see typical multiscale architectures in Figure~\ref{fig:architecture_comparison}, where GXN is well distinguished because intermediate feature interchanging across scales forms a crossing shape.

{\emph{Remark}:} In each individual scale, graph U-net~\cite{ICML2019_Gao} simply uses skip connections while GXN uses multiple graph propagation layers to extract features. The proposed feature-crossing layer is used to fuse intermediate features and cannot be directly applied to graph U-net.

To test our methods, we conduct extensive experiments on several standard datasets for both graph classification and vertex classification. 
Compared to state-of-the-art methods for these two tasks, GXN improves the average classification accuracies by $2.12\%$ and $1.15\%$, respectively. Meanwhile, based on the same model architecture, our VIPool consistently outperforms previous graph pooling methods; and more intermediate connection leads to a better performance. 
\footnote{
The code could be downloaded at \url{https://github.com/limaosen0/GXN}}

% \vspace{-3mm}
\section{Related Works}  
\label{sec:RelatedWorks}
\vspace{-3mm}
\mypar{Multiscale graph neural networks with graph pooling} 
To comprehensively learn the multiscale graph representations, various multiscale network structures have been explored. Hierarchical encoder-decoder structures~\cite{AAAI1816273,DBLP:journals/corr/abs-1802-09612,ICLR2020_Deng} learn graph features just from much coarse scales. LancozsNet~\cite{liao2018lanczosnet} designs various graph filters on the multiscale graphs. Graph U-net~\cite{ICML2019_Gao} and readout functions~\cite{DBLP:journals/corr/abs-1902-06684,ICML2019_Lee} design pyramid structures with skip-connections and combines features from all scales in the last layer. Compared to previous works, the proposed GXN has two main differences. 1) Besides the common late fusion of features, GXN uses intermediate fusion across scales, where the features at various scales in each network layer are fused to embed richer multiscale information. 2) GXN extracts hierarchical multiscale features through a deep network, previous Graph U-net~\cite{ICML2019_Gao} extracts features only once in each scale and then uses skip-connections to fuse feature across scales.

To compress a graph into multiple coarser scales, various methods of graph pooling are proposed. Early  graph pooling methods are usually designed based on graph sampling theory~\cite{ChenVSK:15} or graph coarsening~\cite{SafroS:14}. With the study of deep learning, some works down-scale graphs in data-driven manner. The graph-coarsening-based pooling methods~\cite{4302760,NIPS2016_6081,Simonovsky_2017_CVPR,ijcai2018-490,DBLP:journals/corr/abs-1802-09612,NIPS2018_7729,YuanJ:20} cluster vertices and merge each cluster to a coarsened vertex; however, there is not vertex-to-vertex association to preserve the original vertex information. The vertex-selection-based pooling methods~\cite{ICML2019_Gao,ICML2019_Lee} preserve selected vertices based on their importance, but tend to loss the original graph structures. Compared to previous works, the proposed VIPool in GXN is trained given an explicit optimization for vertex selection, and the pooled graph effectively abstracts the original graph structure.

\mypar{Mutual information estimation and maximization}
Given two variables, to estimate their mutual information whose extact value is hard to compute, some models are constructed based on the parameterization of neural networks. \cite{pmlr-v80-belghazi18a} leverages trainable networks to depict a lower bound of mutual information, which could be optimized toward a precise mutual information estimation. \cite{hjelm2019learning} maximizes the pixel-image mutual information to promote to capture the most informative image patterns via self-supervision. \cite{velickovic2019deep} maximizes the mutual information between a graph and each single vertex, where the representative vertex features are obtained. Similarly, \cite{Sun2020InfoGraph} applies the mutual information maximization on graph classification. Compared to these mutual-information-based studies, the proposed VIPool, which also leverages mutual information maximization on graphs, aims to obtain an optimization for vertex selection by finding the vertices that maximally represent their local neighborhood. We also note that, in VIPool, the data distribution is defined on a single graph, while previous works~\cite{velickovic2019deep,Sun2020InfoGraph} assume to train on the distribution of multiple graphs.

% \vspace{-3mm}
\section{Vertex Infomax Pooling}  
\label{sec:VIPool}
\vspace{-1mm}
Before introducing the overall model, we first propose a new graph pooling method to create multiple scales of a graph. In this graph pooling, we select and preserve a ratio of vertices and connect them based on the original graph structure. Since downscaling graphs would lose information, it is critical to preserve as much information as possible in the pooled graph, which could maximally represent the original graphs. To this end, we propose a novel~\emph{vertex infomax pooling} (VIPool), preserving the vertices that carry high mutual information with their surrounding neighborhoods by mutual information estimation and maximization. The preserved vertices well represent local subgraphs, and they also abstract the overall graph structure based on a vertex selection criterion.

Mathematically, let $G(\mathcal{V}, \mathbf{A})$  be a graph with a set of vertices $\mathcal{V}=\{v_1,\dots,v_N\}$ whose features are $\mathbf{X} = [ \mathbf{x}_1 \cdots \mathbf{x}_N ]^{\top} \in \mathbb{R}^{N \times d}$, and an adjacency matrix $\mathbf{A} \in \{0,1\}^{N \times N}$. We aim to select a subset $\Omega \subset \mathcal{V}$ that contains $|\Omega| = K$ vertices. Considering a criterion function $C(\cdot)$ to quantify the information of a vertex subset, we find the most informative subset through solving the problem,
\begin{equation}
\label{eq:problem}
    \max_{\Omega \subset \mathcal{V}}~~~ C \left( \Omega \right),~~~{\rm subject~to~~~} |\Omega| \ = \ K.
\end{equation}
We design $C(\Omega)$ based on the mutual information between vertices and their corresponding neighborhoods, reflecting the vertices' abilities to express neighborhoods. In the following, we first introduce the computation of vertex-neighborhood mutual information, leading to the definition of $C(\cdot)$; we next select a vertex set by solving~\eqref{eq:problem}; we finally pool a fine graph based on the selected vertices.
 
\begin{figure}[t]
\centering
\includegraphics[width=0.9\textwidth]{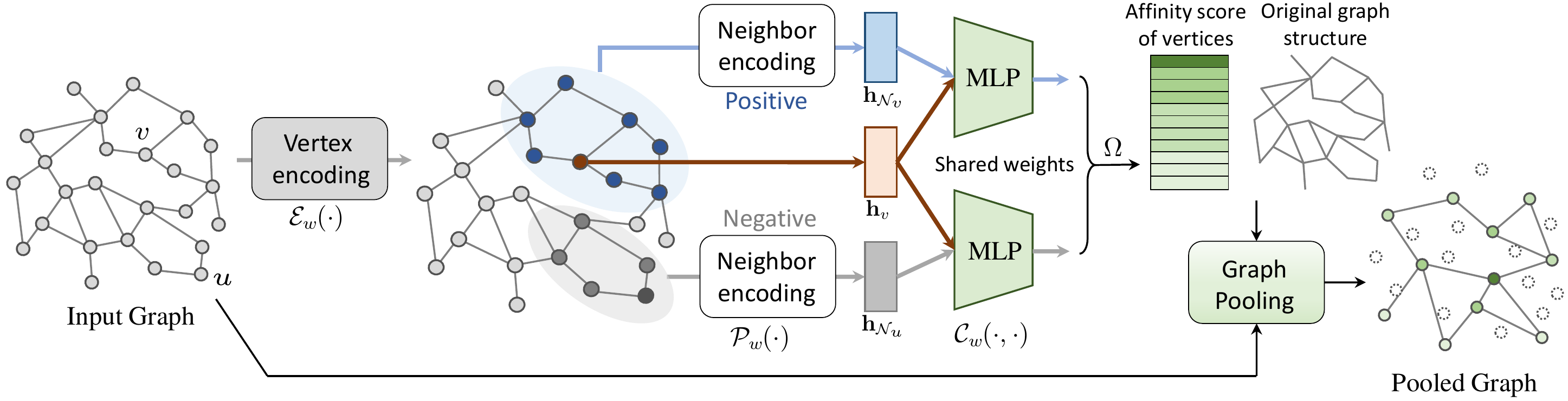}
\vspace{-10pt}
\caption{\small Vertex infomax pooling (VIPool). We evaluate the importance of a vertex based on how much it can reflect its own neighborhood and how much it can discriminate from  an arbitrary neighborhood.}
\vspace{-15pt}
\label{fig:pooling}
\end{figure}

\textbf{Mutual information neural estimation.} 
In a graph $G(\mathcal{V}, \mathbf{A})$, for any selected vertex $v$ in $\Omega \subset \mathcal{V}$, we define $v$'s neighborhood as $\mathcal{N}_v$, which is the subgraph containing the vertices in $\mathcal{V}$ whose geodesic distances to $v$ are no greater than a threshold $R$ according to the original $G(\mathcal{V}, \mathbf{A})$,
 i.e. $\mathcal{N}_v = G(\{ u \}_{d(u, v)\le R}, \mathbf{A}_{\{u\}, \{u\}})$. 
Let a random variable $\mathbf{v}$ be the feature of a randomly picked vertex in $\Omega$, the distribution of ${\bf v}$ is $P_{\mathbf{v}} = P(\mathbf{v} = \mathbf{x}_v)$, where $\mathbf{x}_v$ is the outcome feature value when we pick vertex $v$. 
%Assume the set of vertex features $\{ \mathbf{x}_v\}, \forall v \in \Omega$ follows a certain distribution, noted as $P_{\mathbf{v}} = P(\mathbf{x}_v)$. Similarly, denote the feature of the $v$'s neighborhood $\mathcal{N}_v$ as $\mathbf{y}_{\mathcal{N}_v}$, and its distribution is noted as $P_{\mathbf{n}} = P( \mathbf{y}_{\mathcal{N}_v})$.
Similarly, let a random variable $\mathbf{n}$ be the neighborhood feature associated with a randomly picked vertex in $\Omega$, the distribution of  $\mathbf{n}$ is $P_{\mathbf{n}} = P( \mathbf{n} = \mathbf{y}_{\mathcal{N}_u})$, where $\mathbf{y}_{\mathcal{N}_u}$ 
%= [\mathbf{A}_{\mathcal{N}_u, \mathcal{N}_u}, \{ \mathbf{x}_\nu \}_{\nu \in \mathcal{N}_u}]$ 
is the outcome feature value when we pick vertex $u$'s neighborhood.
%, including both the connectivity information and  vertex features in $\mathcal{N}_u$.
% The joint distribution of vertex and neighborhood features is $P_{\mathbf{v}, \mathbf{n}} = P( \mathbf{v} = \mathbf{x}_v, \mathbf{n} = \mathbf{y}_{\mathcal{N}_v})$, reflecting we randomly select a vertex $v$ and its corresponding neighborhood together. 
The mutual information between selected vertices and neighborhoods is the KL-divergence between the joint distribution $P_{\mathbf{v},\mathbf{n}} = P(\mathbf{v}= \mathbf{x}_v,  \mathbf{n}=\mathbf{y}_{\mathcal{N}_v})$ and the product of marginal distributions $P_{\mathbf{v}} \otimes P_{\mathbf{n}}$:
\begin{eqnarray}
    I^{(\Omega)} \left( \mathbf{v}, \mathbf{n} \right)
    & = & D_{\rm KL} \left( P_{\mathbf{v}, \mathbf{n}} || P_{\mathbf{v}} \otimes P_{\mathbf{n}} \right)
    \nonumber \\  \nonumber 
    & \stackrel{(a)}{\geq} & 
    \sup_{T\in\mathcal{T}} \left\{  
    \mathbb{E}_{\mathbf{x}_v, \mathbf{y}_{\mathcal{N}_v}  \sim  P_{\mathbf{v}, \mathbf{n}}} \left[ T( \mathbf{x}_v,  \mathbf{y}_{\mathcal{N}_v} ) \right] 
    - \mathbb{E}_{ \mathbf{x}_v \sim  P_{\mathbf{v}}, \mathbf{y}_{\mathcal{N}_u} \sim  P_{\mathbf{n}} } \left[ e^{T \left( \mathbf{x}_v,  \mathbf{y}_{\mathcal{N}_u} \right)-1} \right]
    \right\},
\end{eqnarray}
where $(a)$ follows from $f$-divergence representation based on KL divergence~\cite{pmlr-v80-belghazi18a};  $T\in\mathcal{T}$ is an arbitrary function that maps features of a pair of vertex and neighborhood to a real value, here reflecting the dependency of two features. To achieve more flexibility and convenience in optimization, $f$-divergence representation based on a non-KL divergence can be adopted~\cite{NIPS2016_6066}, which still measures the vertex-neighborhood dependency. Here we consider a GAN-like divergence.
\begin{equation*}
\label{eq:GAN}
    I^{(\Omega)}_{\rm GAN} \left( \mathbf{v}, \mathbf{n} \right)
    \ \geq \
    \sup_{T\in\mathcal{T}} \left\{  
    \mathbb{E}_{P_{\mathbf{v}, \mathbf{n}}} 
    \left[ \log \sigma \left( T( \mathbf{x}_v, \mathbf{y}_{\mathcal{N}_v} ) \right) \right] 
    +  \mathbb{E}_{P_{\mathbf{v}}, P_{\mathbf{n}} } \left[ \log \left( 1 - \sigma \left( T \left( \mathbf{x}_v, \mathbf{y}_{\mathcal{N}_u}\right) \right) \right) \right]
    \right\},
\end{equation*}
where $\sigma(\cdot)$ is the sigmoid function. In practice, we cannot go over the entire functional space $\mathcal{T}$ to evaluate the exact value of $I^{(\Omega)}_{\rm GAN}$. Instead, we parameterize $T(\cdot,\cdot)$ by a neural network $T_w(\cdot,\cdot)$, where the subscript $w$ denotes the trainable parameters. Through optimizing over $w$, we obtain a neural estimation of the GAN-based mutual information, denoted as $\widehat{I}^{(\Omega)}_{\rm GAN}$. We can define our vertex selection criterion function to be this neural estimation; that is,
\begin{equation*}
\label{eq:Q}
C \left( \Omega \right)  \ = \ 
\widehat{I}^{(\Omega)}_{\rm GAN} \ =\ 
\max_{w}  \frac{1}{|\Omega|} 
\sum_{v \in \Omega}  
\log \sigma \left( T_w(   {\bf x}_v, {\bf y}_{\mathcal{N}_v}) ) \right) +
\frac{1}{|\Omega|^2} 
\sum_{(v,u) \in  \Omega }  
\log  \big( 1 - \sigma \left(T_w({\bf x}_v, {\bf y}_{\mathcal{N}_u} )) \right)  \big).
\end{equation*}
In $C(\Omega)$, the first term reflects the affinities between vertices and their own neighborhoods; and the second term reflects the differences between vertices and arbitrary neighborhoods. Notably, a higher $C$ score indicates that  vertices maximally reflect their own neighborhoods and meanwhile minimally reflect arbitrary neighborhoods. To specify $T_w$, we consider $T_w \left(  \mathbf{x}_v, \mathbf{y}_{\mathcal{N}_u} \right)  = \mathcal{S}_w  \left( \mathcal{E}_w(\mathbf{x}_v), \mathcal{P}_w(\mathbf{y}_{\mathcal{N}_u} ) \right)$, where the subscript $w$ indicates the associated functions are trainable\footnote{The trainable parameters in $\mathcal{E}_w(\cdot)$,  $\mathcal{P}_w(\cdot)$, and $\mathcal{S}_w(\cdot, \cdot)$ are not weight-shared.}, $\mathcal{E}_w(\cdot)$ and $\mathcal{P}_w(\cdot)$ are embedding functions of vertices and neighborhoods, respectively, and $\mathcal{S}_w(\cdot, \cdot)$ is an affinity function to quantify the affinity between vertices and neighborhoods; see an illustration in  Figure~\ref{fig:pooling}. We implement $\mathcal{E}_w(\cdot)$ and $\mathcal{S}_w(\cdot, \cdot)$ by multi-layer perceptrons (MLPs), and implement $\mathcal{P}_w(\cdot)$ by aggregating vertex features and neighborhood connectivities in ${\bf y}_{\mathcal{N}_u}$; that is
\begin{equation}
\label{eq:pw}
\mathcal{P}_w({\bf y}_{\mathcal{N}_u}) ~=~
\frac{1}{R}
\sum_{r=0}^{R}
\sum_{\nu\in\mathcal{N}_u}
\left( (\widetilde{\bf D}^{-1/2}\widetilde{\bf A}\widetilde{\bf D}^{-1/2})^r
\right)_{\nu,u}
{\bf W}^{(r)}
\mathcal{E}_w({\bf x}_{\nu}),
\ \ \ \ 
\forall u \in \Omega,
\end{equation}
where $\widetilde{\bf A} = {\bf A} + {\bf I}\in\{0,1\}^{N \times N}$ denotes the self-connected graph adjacency matrix and $\widetilde{\bf D}$ is the degree matrix of $\widetilde{\bf A}$; ${\bf W}^{(r)}$ is the trainable weight associated with the $r$th hop of neighbors; $\mathcal{P}_w(\cdot)$.
The detailed derivation is presented in Appendix A.

When we maximize $C(\Omega)$ by training $\mathcal{E}_w(\cdot)$,  $\mathcal{P}_w(\cdot)$ and  $\mathcal{S}_w(\cdot, \cdot)$, we estimate the mutual information between vertices in $\Omega$ and their neighborhoods. {This is similar to deep graph infomax (DGI)~\cite{velickovic2019deep},  which estimates the mutual information between any vertex feature and a global graph embedding. Both DGI and the proposed VIPool apply the techniques of mutual information neural estimation~\cite{pmlr-v80-belghazi18a,hjelm2019learning} to the graph domain; however, there are two major differences. First, DGI aims to train a graph embedding function while VIPool aims to evaluate the importance of a vertex via its affinity to its neighborhood. Second, DGI considers the relationship between a vertex and an entire graph while VIPool learns the dependency between a vertex and a neighborhood. By varying the neighbor-hop $R$ of $\mathcal{N}_u$ in Eq.~\eqref{eq:pw}, VIPool is able to tradeoff local and global information.}

\textbf{Solutions for vertex selection.}
To solve the vertex selection problem~\eqref{eq:problem}, we consider the submodularity of mutual information~\cite{ChenHHKK:15} and employ a greedy algorithm: we select the first vertex with maximum $C(\Omega)$ with $|\Omega|=1$; and we next add a new vertex sequentially by maximizing $C(\Omega)$ greedily; {however, it is computationally expensive to evaluate $C(\Omega)$ for two reasons: (i) for any vertex set $\Omega$, we need to solve an individual optimization problem; and (ii) the second term of $C(\Omega)$ includes all the pairwise interactions involved with quadratic computational cost. To address issue (i), we set the vertex set to all the vertices in the graph, maximize $\widehat{I}^{(\mathcal{V})}_{\rm GAN}$ to train $\mathcal{E}_w(\cdot)$, $\mathcal{P}_w(\cdot)$ and $\mathcal{S}_w(\cdot, \cdot)$. We then fix those three functions and evaluate $\widehat{I}^{(\Omega)}_{\rm GAN}$. To address issue (ii), we perform negative sampling, approximating the second term~\cite{MikolovSCCD:13}, where we sample negative neighborhoods $\mathcal{N}_u$ from the entire graph, whose number equals the number of positive vertex samples; that is, $|\Omega|$.}

\textbf{Graph pooling and unpooling.}
After solving problem.~\eqref{eq:problem}, we obtain $\Omega$ that contains $K$ unique vertices selected from $\mathcal{V}$. To implement \emph{graph pooling}, we further consider the distinct importance of different vertices in $\Omega$, we compute an affinity score for each vertex based on its ability to describe its neighborhood. For vertex $v$ with feature ${\bf x}_v$ and neighborhood feature ${\bf y}_{\mathcal{N}_v}$, the affinity score is
\begin{equation}
   \label{eq:weight_score}
    {a}_v ~=~ 
    \sigma
    \left(\mathcal{S}_w(
    \mathcal{E}_w({\bf x}_{v}), 
    \mathcal{P}_w({\bf y}_{\mathcal{N}_v}))
    \right) \in [0,1], 
    \ \ \ \ \forall v\in\Omega.
\end{equation}
Eq.~\eqref{eq:weight_score} considers the affinity only between a vertex and its own neighborhood, showing the degree of vertex-neighborhood information dependency.  We collect ${a}_v$ for $\forall v\in\Omega$ to form an affinity vector ${\bf a}\in[0,1]^{K}$. For {graph data pooling}, the pooled vertex feature ${\bf X}_{\Omega}=\mathbf{X}({\rm id},:)\odot (\mathbf{a1}^{\top})\in\mathbb{R}^{K \times d}$, where ${\rm id}$ denotes selected vertices's indices that are originally in $\mathcal{V}$, ${\bf 1}$ is an all-one vector, and $\odot$ denotes the element-wise multiplication. With the affinity vector ${\bf a}$, we assign an importance to each vertex and provide a path for back-propagation to flow gradients. As for {graph structure pooling}, we calculate ${\bf A}_{\Omega}= {\rm Pool}_{A}({\bf A})$, and we consider three approaches to implement ${\rm Pool}_{A}(\cdot)$:

$\bullet$ Edge removal, i.e. ${\bf A}_{\Omega}={\bf A}({\rm id}, {\rm id})$. This is simple, but loses significant structural information;

\vspace{-0.8mm}
$\bullet$ Kron reduction~\cite{DorflerB:13}, which is the Schur complement of the graph Laplacian matrix and preserves the graph spectral properties, but it is computationally expensive  due to the matrix inversion;

\vspace{-0.8mm}
$\bullet$ Cluster-connection, i.e. ${\bf A}_{\Omega} = {\bf S}{\bf A}{\bf S}^{\top}$ with ${\bf S}={\rm softmax}({\bf A}({\rm id},:))\in [0,1]^{K \times N}$. Each row of ${\bf S}$ represents the neighborhood of a selected vertex and the softmax function is applied for normalization. The intuition is to merge the neighboring information to the selected vertices~\cite{NIPS2018_7729}.

Cluster-connection is our default implementation of the graph structure pooling. Figure~\ref{fig:pooling} illustrates the overall process of vertex selection and graph pooling process.

To implement \emph{graph unpooling}, inspired by~\cite{ICML2019_Gao}, we design an inverse process against graph pooling. We initialize a zero matrix for the unpooled graph data, $ {\bf X}' = {\bf O}\in\{0\}^{N \times d}$; and then, fill it by fetching the vertex features according to the original indices of retrained vertices; that is, ${\bf X}'({\rm id},:)={\bf X}_{\Omega}$. We then interpolate it through a graph propagation layer (implemented by graph convolution~\cite{ICLR2017_Kipf}) to propagate information from the vertices in $\Omega$ to the padded ones via the original graph structure.

% \vspace{-3mm}
\section{Graph Cross Network}
\begin{figure}[t]
\centering
\includegraphics[width=0.9\textwidth]{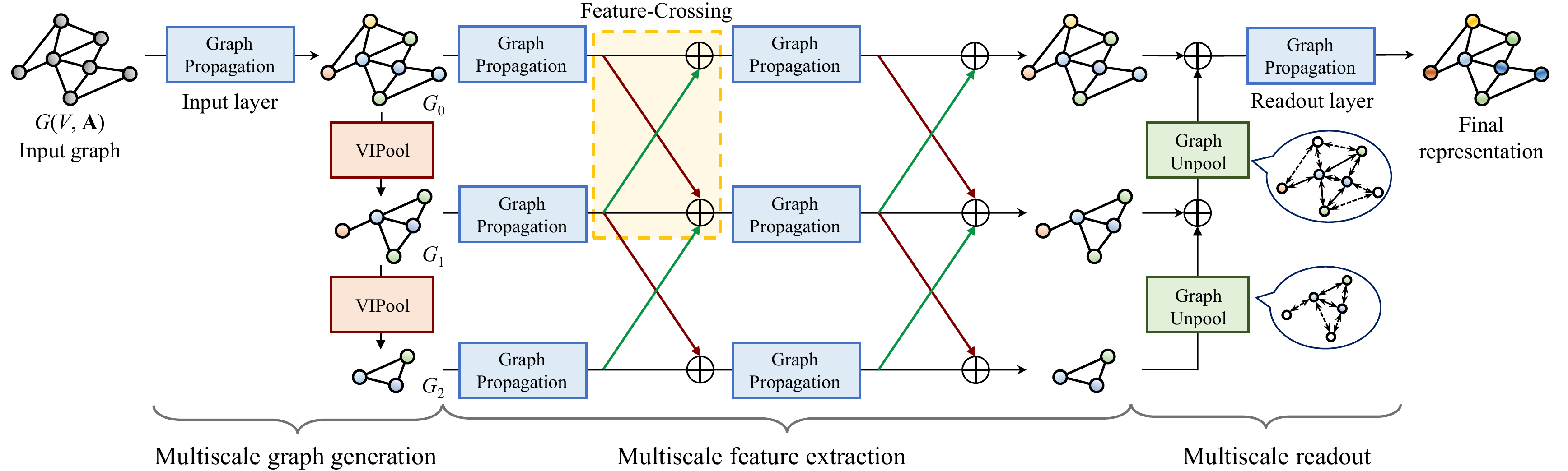}
\vspace{-5pt}
\caption{\small GXN architecture. We show an exemplar model with 3 scales and 4 feature-crossing layers.}
\label{fig:architecture}
\vspace{-14pt}
\end{figure}

\vspace{-1mm}
In this section, we propose the architecture of our~\emph{graph cross network} (GXN) for multiscale graph feature learning; see an exemplar model with $3$ scales and $4$ feature-crossing layers in Figure~\ref{fig:architecture}.  The graph pooling/unpooling operations apply VIPool proposed in Section~\ref{sec:VIPool} and the graph propagation layers adopt the graph convolution layers~\cite{ICLR2017_Kipf}. A key ingredient of GXN is that we design~\emph{feature-crossing layers} to enhance multiscale information fusion. The entire GXN includes three stages: multiscale graphs generation, multiscale features extraction
and multiscale readout.
%1)~\emph{multiscale graphs generation}, where GXN automatically generates nested graphs with graph pooling; 2)~\emph{multiscale features extraction}, where GXN exploits comprehensive multiscale deep graph features; and 3)~\emph{multiscale readout}, where GXN  fuses the multiscale features to obtain final features.

{\bf Multiscale graphs generation.} 
Given an input graph $G(\mathcal{V},{\bf A})$ with vertex features, $\mathbf{X}\in\mathbb{R}^{N \times d}$, we aim to create graph representations at multiple scales. 
We first employ a graph propagation layer on the input graph to initially embed the finest scale of graph as $G_0(\mathcal{V}_0, \mathbf{A}_{0})$ with $\mathcal{V}_0 = \mathcal{V}$, $\mathbf{A}_{0} = {\bf A}$ and vertex representations $\mathbf{X}_{0}$, where the graph propagation layer is implemented by a graph convolution layer~\cite{ICLR2017_Kipf}. We then recursively apply VIPool for $S$ times to obtain a series of coarser scales of graph $G_1(\mathcal{V}_1, \mathbf{A}_1), \dots, G_S(\mathcal{V}_S, \mathbf{A}_S)$ 
and corresponding vertex features $\mathbf{X}_1, \dots, \mathbf{X}_S$ from $G_0$ and $\mathbf{X}_0$, respectively, where $|\mathcal{V}_{s}|>|\mathcal{V}_{s'}|$ for $\forall~ 1 \leq s < s' \leq S$.

% employ a input graph propagation layer that takes  ${\bf A}$ and $\mathbf{X}$ as inputs to preliminarily update vertex information with graph structures and obtain an embedded graph $G_0(\mathcal{V}_0, \mathbf{A}_{0})$ with $\mathcal{V}_{0}$ containing $N$ vertices whose features are $\mathbf{X}_{0}\in\mathbb{R}^{N \times d_{\rm h}}$ and $\mathbf{A}_0 = \mathbf{A}$. The graph propagation layer can be modeled by a graph convolution layer; that is  $ {\bf X}_0 ~=~ {\rm ReLU}(\widetilde{{\bf D}}^{-1/2} \widetilde{{\bf A}} \widetilde{{\bf D}}^{-1/2}{\bf X}{\bf W}_{\rm in}), $ where $\widetilde{{\bf A}}={\bf A}+{\bf I}$ is the self-connected adjacency matrix, whose degree matrix is $\widetilde{{\bf D}}$; ${\bf W}_{\rm in}\in\mathbb{R}^{d\times d_{\rm h}}$ is the trainable weights of the input layer; ${\rm ReLU}(\cdot)$ is the ReLU activation.

{\bf Multiscale features extraction.}
Given multiscale graphs, we build a graph neural network at each scale to extract features. Each  network consists of a sequence of graph propagation layers. To further enhance the information flow across scales, we propose feature-crossing layers between two consecutive scales at various network layers, which allows multiscale features to communicate and merge in the intermediate layers. Mathematically, at scale $s$ and the any network layer, let the feature matrix of graph $G_s$ be ${\bf X}_{{\rm h},s}$, the feature of graph $G_{s-1}$ after pooling be ${\bf X}_{{\rm h},s}^{({\rm p})}$, and the feature of graph $G_{s+1}$ after unpooling be ${\bf X}_{{\rm h},s}^{({\rm up})}$, the obtained vertex feature ${\bf X}_{{\rm h},s}'$ is formulated as
\vspace{-2mm}
\begin{equation*}
  {\bf X}_{{\rm h},s}' = {\bf X}_{{\rm h},s} + {\bf X}_{{\rm h}, s}^{({\rm p})} + {\bf X}_{{\rm h}, s}^{({\rm up})}, ~~~~ 0<s<S.
 \end{equation*}
% To design a feature-crossing layer, we use graph pooling on a finer scale and graph unpooling on a coarser scale to convert the corresponding vertex features to the other scale, and then add converted features with original features in each scale to achieve fusion; see Figure~\ref{fig:architecture}. 
%Mathematically, at scale $s$ and the h-th layer, let the feature matrix of graph $G_s$ be ${\bf X}_{{\rm h},s}$, the feature of graph $G_{s-1}$ after pooling be ${\bf X}_{{\rm h},s-1}^{({\rm c})}$, and the feature of graph $G_{s+1}$ after unpooling be ${\bf X}_{{\rm h},s+1}^{({\rm f})}$, the obtained vertex feature ${\bf X}_{{\rm h},s}'$ is formulated as
% \begin{equation*}
%   {\bf X}_{{\rm h},s}' =
%   \begin{cases}
%   {\bf X}_{{\rm h},s}+ {\bf X}_{{\rm h},s+1}^{({\rm f})}, &s=0 \\
%   {\bf X}_{{\rm h},s} + {\bf X}_{{\rm h}, s-1}^{({\rm c})} + {\bf X}_{{\rm h}, s+1}^{({\rm f})}, &0<s<S  \\
%   {\bf X}_{{\rm h},s} + {\bf X}_{{\rm h},s-1}^{({\rm c})}. &s=S
%   \end{cases}
%   \end{equation*}
For $s=0$ or $S$, ${\bf X}_{{\rm h},s}$ is not fused by features from finer or coarser scales; see Figure~\ref{fig:architecture}. 
The graph pooling/unpooling here uses the same vertices as those obtained in multiscale graph generation to associate the vertices at different layers, but the affinity score ${\bf a}$ in each feature-crossing layer is trained independently to reflect the vertex importance at different levels. Note that the vertex-to-vertex association across scales is important here for feature-crossing and VIPool nicely fits it.

% {\HC In graph pooling and unpooling, we use the same vertex indices as obtained during multiscale graph generation to match the vertex identities in different layers. In addition, we use the current hidden features to calculate their own importance scores for feature converting at various feature-crossing layers. The individual importance scores flexibly represent the vertex representativeness in different feature levels.}

{\bf Multiscale readout.}
After multiscale feature extraction, we combine deep features at all the scales together to obtain the final representation. To align features at different scales, we adopt a sequence of graph unpooling operations implemented in the VIPool to transform all features to the original scale; see Figure~\ref{fig:architecture}.
% \begin{equation}
%     {\bf H}_0' = {\bf H}_0 + \lambda(\sum_{s=1}^{S}{\bf H}_{s}^{({\rm f})})
% \end{equation}
We finally leverage a readout graph propagation layer to further embed the fused multiscale features and generate the readout graph representation for various downstream tasks. In this work, we consider both graph classification and vertex classification.

\mypar{Model Training}
To train GXN, we consider the training loss with two terms: a graph-pooling loss $\mathcal{L}_{\rm pool}= - \widehat{I}^{(\mathcal{V})}_{\rm GAN}$ and a task-driven loss $\mathcal{L}_{\rm task}$. For graph classification, the task-driven loss is the cross-entropy loss between the predicted and ground-truth graph labels, $\mathcal{L}_{\rm task}=-{\bf y}^{\top}\log(\hat{\bf y})$, where ${\bf y}$ and $\hat{\bf y}$ are ground-truth label and predicted label of a graph; for vertex classification, it is the cross-entropy loss between the predictions and ground-truth vertex labels, $\mathcal{L}_{\rm task}=-\sum_{v\in\mathcal{V}_L}{\bf y}_v^{\top}\log(\hat{\bf y}_v)$, where ${\bf y}_v$ and $\hat{\bf y}_v$ are ground-truth and predicted vertex labels, and $\mathcal{V}_L$ contains labeled vertices. We finally define the overall loss as
$
\mathcal{L} ~=~ \mathcal{L}_{\rm task} + \alpha \mathcal{L}_{\rm pool},
$
where the hyper-parameter $\alpha$ linearly decays per epoch from 2 to 0 during training, balancing a final task and vertex pooling\footnote{ VIPool is trained through both $\mathcal{L}_{\rm pool}$ and $\mathcal{L}_{\rm task}$, which makes graph pooling  adapt to a specific task.}.

% \begin{algorithm}[htb] 
% \caption{Training the GXN.} 
% \label{alg:Framwork} 
% \begin{algorithmic}[1] 
% \Require 
% The input graph $G(\mathcal{V},{\bf A})$ with vertex features ${\bf X}$; 
% The number of scales $S$;
% The percentage of preserveed vertex in each scale $k_s$.
% \Ensure 
% Multiscale graph representations and results of graph-based tasks. 
% \Repeat
% \State \textbf{Training-Forward:}
% \State Determine $\Omega$ for coarser scales of graphs by solving problem Eq.~\eqref{eq:problem} and perform graph pooling.
% \State Run the rest modules of GXN for multiscale graph representation learning.
% \State Calculate $\mathcal{L}$ for a specific downstream task.
% \State \textbf{Training-Backward:}
% \State Based on $\mathcal{L}$, update the model parameters using stochastic gradient descent.
% \Until{$\mathcal{L}$ converges.}
% \State {\bf Model Test.} Run the feed-forward GXN model with the test data.
% \end{algorithmic} 
% \end{algorithm}

%Different from recent works which use self-supervised paradigms to estimate and maximize the mutual information~\cite{hjelm2019learning,velickovic2019deep,Sun2020InfoGraph}, {\HC TBD}

\section{Experimental Results}
\vspace{-1mm}
\subsection{Datasets and Experiment Setup}
\vspace{-2mm}
\textbf{Datasets.} To test our GXN, we conduct extensive experiments for graph classification and vertex classification on several datasets. 
For {\bf graph classification}, we use social network datasets: IMDB-B, IMDB-M and COLLAB~\cite{NIPS2015_5880}, and bioinformatic datasets: D\&D~\cite{DD}, PROTEINS~\cite{NIPS2013_5155}, and ENZYMES~\cite{ENZYMES}. Table~\ref{tab:graph_classification} shows the dataset information. Note that no vertex feature is provided in three social network datasets, and we use one-hot vectors to encode the vertex degrees as vertex features, explicitly utilizing some structural information. We use the same dataset separation as in~\cite{ICML2019_Gao}, perform 10-fold cross-validation, and show the average accuracy for evaluation.
For {\bf vertex classification}, we use three classical citation networks: Cora, Citeseer and Pubmed~\cite{ICLR2017_Kipf}. We perform both full-supervised and semi-supervised vertex classification; that is, for full-supervised classification, we label all the vertices in training sets for model training, while for semi-supervised, we only label a few vertices (around 7\% on average) in training sets. We use the default separations of training/validation/test subsets. See more information of all used datasets in Appendix.

\textbf{Model configuration.} 
We implement GXN with PyTorch 1.0 on one GTX-1080Ti GPU.
For {\bf graph classification}, we consider three scales, which preserve $50\%$ to $100\%$ vertices from the original scales, respectively. For both input and readout layers, we use 1-layer GCNs; for multiscale feature extraction, we use $2$ GCN layers followed by ReLUs at each scale and feature-crossing layers between any two consecutive scales at any layers. After the readout layers, we unify the embeddings of various graphs to the same dimension by using the same SortPool in DGCNN~\cite{AAAI1817146}, AttPool~\cite{Huang_2019_ICCV} and Graph U-Net~\cite{ICML2019_Gao}.
In the VIPool, we use a $2$-layer MLP and $R$-layer GCN ($R=1$ or $2$) as $\mathcal{E}_w(\cdot)$ and $\mathcal{P}_w(\cdot)$, and use a linear layer as $\mathcal{S}_w(\cdot, \cdot)$. The hidden dimensions are $48$.
To improve the efficiency of solving problem~\eqref{eq:problem}, we modify $C(\Omega)$ by preserving only the first term. In this way, we effectively reduce the computational costs to sovle~\eqref{eq:problem} from $\mathcal{O}(|V|^2)$ to $\mathcal{O}(|V|)$, and each vertex contributes the vertex set independently. The optimal solution is {\bf top-$K$} vertices. We compare the outcomes of $C(\Omega)$ by the greedy algorithm and top-k method in Figure~\ref{fig:Qcompare}.
For {\bf vertex classification}, we use similar architecture as in graph classification, while the hidden feature are 128-dimension. We directly use the readout layer for vertex classification. In the loss function $\mathcal{L}$, $\alpha$ decays from $2$ to $0$ during training, where the VIPool needs fast convergence for vertex selection; and the model gradually focuses more on tasks based on the effective VIPool. We use Adam optimizer~\cite{Adam} and the learining rates range from 0.0001 to 0.001 for different datasets.
\vspace{-2mm}
\subsection{Comparison}
\vspace{-2mm}
\textbf{Graph classification.} We compare the proposed GXN to representative GNN-based methods, including 
PatchySAN~\cite{pmlr-v70-Niepert16}, 
ECC~\cite{Simonovsky_2017_CVPR}, 
Set2Set~\cite{pmlr-v70-gilmer17a}, 
DGCNN~\cite{AAAI1817146}, 
DiffPool~\cite{NIPS2018_7729}, 
Graph U-Net~\cite{ICML2019_Gao}, 
SAGPool~\cite{ICML2019_Lee}, 
AttPool~\cite{Huang_2019_ICCV}, 
and StructPool~\cite{YuanJ:20},
where most of them performed multiscale graph feature learning. We unify the train/test data splits and processes of model selection for fair comparison~\cite{Errica2020A}.
Additionally, we design several variants of GXN: 1) to test the superiority of VIPool, we apply gPool~\cite{ICML2019_Gao}, SAGPool~\cite{ICML2019_Lee} and AttPool~\cite{Huang_2019_ICCV} in the same architecture of GXN, denoted as `GXN (gPool)', `GXN (SAGPool)' and `GXN (AttPool)', respectively; 2) we investigate different feature-crossing mechanism, including various crossing directions and crossing positions. Table~\ref{tab:graph_classification} compares the accuracies of various methods for graph classification.
\begin{table}[t]
  \centering
  \scriptsize
  \caption{Graph classification accuracies (\%) of different methods on different datasets. GXN (gPool) and GXN (SAGPool) denote that we apply previous pooling operations, gPool~\cite{ICML2019_Gao} and SAGPool~\cite{ICML2019_Lee} in our GXN framework, respectively. Various fashions of feature-crossing are presented, including fusion of coarse-to-fine ($\uparrow$), fine-to-coarse ($\downarrow$), no feature-crossing (noCross), and feature-crossing at early, late and all layers of networks.}
  \setlength{\tabcolsep}{3.46mm}{
  \begin{tabular}{ccccccc}
  \specialrule{0.08em}{0pt}{1pt}
  Dataset & IMDB-B & IMDB-M & COLLAB & D\&D & PROTEINS & ENZYMES \\
  % \specialrule{0.05em}{1pt}{1pt}
  \# Graphs (Classes) & 1000 (2) & 1500 (3) & 5000 (3) & 1178 (2) & 1113 (2) & 600 (6) \\
%   \# Classes & 2 & 2 & 3 & 3 & 2 & 2 & 6 \\
  Avg. \# Vertices & 19.77 & 13.00 & 74.49 & 284.32 & 39.06 & 32.63 \\
  \specialrule{0.05em}{1pt}{1pt}
  PatchySAN~\cite{pmlr-v70-Niepert16} & 76.27 $\pm$ 2.6 & 69.70 $\pm$ 2.2 & 43.33 $\pm$ 2.8 & 72.60 $\pm$ 2.2 & 75.00 $\pm$ 2.8 & - \\
  ECC~\cite{Simonovsky_2017_CVPR} & 67.70 $\pm$ 2.8 & 43.48 $\pm$ 3.0 & 67.82 $\pm$ 2.4 & 72.57 $\pm$ 4.1 & 72.33 $\pm$ 3.4 & 29.50 $\pm$ 7.6 \\
  Set2Set~\cite{pmlr-v70-gilmer17a} & - & - & 71.75 & 78.12 & 74.29 & {60.15} \\
  DGCNN~\cite{AAAI1817146} & 69.20 $\pm$ 3.0 & 45.63 $\pm$ 3.4 & 71.22 $\pm$ 1.9 & 76.59 $\pm$ 4.1 & 72.37 $\pm$ 3.4 & 38.83 $\pm$ 5.7 \\
  DiffPool~\cite{NIPS2018_7729} & 68.40 $\pm$ 6.1 & 45.62 $\pm$ 3.4 & 74.83 $\pm$ 2.0 & 75.05 $\pm$ 3.4 & 73.72 $\pm$ 3.5 &  {\bf 61.83 $\pm$ 5.3} \\
  Graph U-Net~\cite{ICML2019_Gao} & 73.40 $\pm$ 3.7 & 50.27 $\pm$ 3.4 & 77.58 $\pm$ 1.6 & 82.14 $\pm$ 3.0 & 77.20 $\pm$ 4.3 & 50.33 $\pm$ 6.3 \\
  SAGPool~\cite{ICML2019_Lee} & 72.80 $\pm$ 2.3 & 49.43 $\pm$ 2.6 & 76.92 $\pm$ 1.6 & 78.35 $\pm$ 3.5 & 78.28 $\pm$ 4.0 & 52.67 $\pm$ 5.8 \\
  AttPool~\cite{Huang_2019_ICCV} & 73.60 $\pm$ 2.4 & 50.67 $\pm$ 2.7 & 77.04 $\pm$ 1.3 & 79.20 $\pm$ 3.8 & 76.50 $\pm$ 4.2 & 55.33 $\pm$ 6.2 \\
%   StructPool~\cite{YuanJ:20} & {74.70} & {52.47} & 74.22 & 84.19 & {80.36} & {\bf 63.83} \\
  \specialrule{0.05em}{0pt}{1pt}
  GXN & {\bf 78.60 $\pm$ 2.3} & {\bf 55.20 $\pm$ 2.5} & {\bf 78.82 $\pm$ 1.4} &  {\bf 82.68 $\pm$ 4.1} & {\bf 79.91 $\pm$ 4.1} & 57.50 $\pm$ 6.1 \\
  GXN (gPool) & 76.40 $\pm$ 2.6 & 53.62 $\pm$ 2.6 & 77.73 $\pm$ 1.1 & 81.94 $\pm$ 4.3 & 78.44 $\pm$ 3.8 & 55.43 $\pm$ 5.3 \\
  GXN (SAGPool) & 76.90 $\pm$ 2.4 & 53.45 $\pm$ 2.8 & 78.10 $\pm$ 1.5 & 82.16 $\pm$ 4.2 & 78.58 $\pm$ 4.1 & 56.18 $\pm$ 6.1 \\
  GXN (AttPool) & 77.30 $\pm$ 2.3 & 54.71 $\pm$ 2.9 & 78.22 $\pm$ 1.2 & 82.43 $\pm$ 3.9 & 78.09 $\pm$ 4.3 & 56.33 $\pm$ 5.8 \\
  \specialrule{0.05em}{0pt}{1pt}
  GXN ($\uparrow$) & 77.10 $\pm$ 2.2 & 54.22 $\pm$ 2.5 & 78.16 $\pm$ 1.2 & 82.03 $\pm$ 3.8 & 78.87 $\pm$ 3.8 & 56.83 $\pm$ 5.8 \\
  GXN ($\downarrow$) & 76.60 $\pm$ 2.1 & 54.08 $\pm$ 2.3 & 77.68 $\pm$ 1.0 & 81.67 $\pm$ 4.0 & 78.64 $\pm$ 4.2 & 55.33 $\pm$ 5.6 \\
%   \specialrule{0.05em}{0pt}{1pt}
  GXN (noCross) & 74.80 $\pm$ 2.1 &  52.68 $\pm$ 2.7 & 76.95 $\pm$ 1.4 &  81.23 $\pm$ 3.9 &  78.26 $\pm$ 3.9 & 55.17 $\pm$ 5.7 \\
  GXN (early) & 77.50 $\pm$ 2.4 & 54.27 $\pm$ 2.6 & 78.18 $\pm$ 1.1 &  82.43 $\pm$ 4.0 &  79.20 $\pm$ 4.0 & 56.67 $\pm$ 5.4 \\
  GXN (late) & 76.40 $\pm$ 2.0 & 53.83 $\pm$ 2.4 & 77.48 $\pm$ 1.5 & 82.16 $\pm$ 3.6 & 79.03 $\pm$ 4.2 & 56.00 $\pm$ 5.9 \\
  GXN (all) & {\bf 78.60 $\pm$ 2.3} & {\bf 55.20 $\pm$ 2.5} & {\bf 78.82 $\pm$ 1.4}  & {\bf 82.68 $\pm$ 4.1} & {\bf 79.91 $\pm$ 4.1} & 57.50 $\pm$ 6.1 \\
  \specialrule{0.08em}{1pt}{1pt}
  \end{tabular}}
  \label{tab:graph_classification}
  \vspace{-10pt}
\end{table}
We see that our model outperforms the state-of-the-art methods on 5 out of 6 datasets, achieving an improvement by $2.12\%$ on average accuracies. Besides, VIPool and more feature-crossing lead to better performance. We also show the qualitative results of vertex selection of different graph pooling methods in Appendix.

% Additionally, we achieve the best results on 4 out of 6 benchmarks and improve the state-of-the-art performance. 
% It can be observed that previous state-of-the-art models usually have varied performances over different datasets.
% Compared with previous models, the AttPool behaves more consistently over various datasets, which shows our proposed affinity mechanism can extract better representation for a wide family of graph structured data.

\textbf{Vertex classification.} We compare GXN to state-of-the-art methods: DeepWalk~\cite{DBLP:journals/corr/PerozziAS14}, GCN~\cite{ICLR2017_Kipf}, GraphSAGE~\cite{NIPS2017_6703}, FastGCN~\cite{chen2018fastgcn}, ASGCN~\cite{NIPS2018_7707}, and Graph U-Net~\cite{ICML2019_Gao} for vertex classification. We reproduce these methods for both full-supervised and semi-supervised learning based on their official codes. Table~\ref{tab:node_classification} compares the vertex classification accuracies of various methods.
\begin{table}[t]
  \centering
  \scriptsize
  \caption{Vertex classification accuracies (\%) of different methods, where `full-sup.' and `semi-sup.' denote the scenarios of full-supervised and semi-supervised vertex classification, respectively.}
  \setlength{\tabcolsep}{3.90mm}{
  \begin{tabular}{c|cc|cc|cc}
  \specialrule{0.08em}{0pt}{1pt}
  Dataset & \multicolumn{2}{c|}{Cora} & \multicolumn{2}{c|}{Citeseer} & \multicolumn{2}{c}{Pubmed} \\
  % \specialrule{0.05em}{1pt}{1pt}
  \# Vertices (Classes) & \multicolumn{2}{c|}{2708 (7)} & \multicolumn{2}{c|}{3327 (6)} & \multicolumn{2}{c}{19717 (3)} \\
  %\specialrule{0.05em}{1pt}{1pt}
  Supervision & full-sup. & semi-sup. & full-sup. & semi-sup. & full-sup. & semi-sup.\\

  \specialrule{0.05em}{1pt}{1pt}
  DeepWalk~\cite{DBLP:journals/corr/PerozziAS14} & 78.4 $\pm$ 1.7 & 67.2 $\pm$ 2.0 & 68.5 $\pm$ 1.8 & 43.2 $\pm$ 1.6 & 79.8 $\pm$ 1.1 & 65.3 $\pm$ 1.1 \\
  ChebNet~\cite{NIPS2016_6081} & 86.4 $\pm$ 0.5 & 81.2 $\pm$ 0.5 & 78.9 $\pm$ 0.4 & 69.8 $\pm$ 0.5 & 88.7 $\pm$ 0.3 & 74.4 $\pm$ 0.4 \\
  GCN~\cite{ICLR2017_Kipf} & 86.6 $\pm$ 0.4 & 81.5 $\pm$ 0.5 & 79.3 $\pm$ 0.5 & 70.3 $\pm$ 0.5 & 90.2 $\pm$ 0.3 & 79.0 $\pm$ 0.3\\
  GAT~\cite{velickovic2018graph} & 87.8 $\pm$ 0.7 & 83.0 $\pm$ 0.7 & 80.2 $\pm$ 0.6 & 73.5 $\pm$ 0.7 & 90.6 $\pm$ 0.4 & 79.0 $\pm$ 0.3 \\
  FastGCN~\cite{chen2018fastgcn} & 85.0 $\pm$ 0.8 & 80.8 $\pm$ 1.0 & 77.6 $\pm$ 0.8 & 69.4 $\pm$ 0.8 & 88.0 $\pm$ 0.6 & 78.5 $\pm$ 0.7 \\
  ASGCN~\cite{NIPS2018_7707} & 87.4 $\pm$ 0.3 & - & 79.6 $\pm$ 0.2 & - & 90.6 $\pm$ 0.3 & -\\
  Graph U-Net~\cite{ICML2019_Gao} & - & 84.4 & - & 73.2 & - & 79.6 \\
  \specialrule{0.05em}{0pt}{1pt}
  GXN & {\bf 88.9 $\pm$ 0.4} & {\bf 85.1 $\pm$ 0.6} & {\bf 80.9 $\pm$ 0.4} & {\bf 74.8 $\pm$ 0.4} & {\bf 91.8 $\pm$ 0.3} & {\bf 80.2 $\pm$ 0.3} \\
  GXN (noCross) & 87.3 $\pm$ 0.4 & 83.2 $\pm$ 0.5 & 79.5 $\pm$ 0.4 & 73.7 $\pm$ 0.3 & 91.1 $\pm$ 0.2 & 79.6 $\pm$ 0.3 \\
  \specialrule{0.08em}{1pt}{1pt}
  \end{tabular}}
  \label{tab:node_classification}
  \vspace{-12pt}
\end{table}
Considering both full-supervised and semi-supervised settings, we see that our model achieves higher average accuracy by $1.15\%$. 
We also test a degraded GXN without any feature-crossing layer, and we see that the feature-crossing layers improves the accuracies by $1.10\%$ on average; see more results in Appendix.
\vspace{-2mm}
\subsection{Model Analysis}
\vspace{-2mm}
We further conduct detailed analysis about the GXN architecture and VIPool.
% We first investigate the proposed VIPool operation, where we analyze different vertex selection algorithms and show the pooling results; we then explore different model architectures to test their performances.

\begin{figure}[!b]
\vspace{-10pt}
  \begin{minipage}[b]{0.49\textwidth}
    \centering
    \setlength{\tabcolsep}{1.9mm}{
    \begin{tabular}{c|c|ccc}
      \hline
      ~ & ~ & \multicolumn{3}{c}{accuracy} \\
      \hline
      ~ & \tabincell{c}{\# cross} & 1 & 2 & 3\\
      \hline
      \multirowcell{5}{\# scales} & 1 & 74.50 & 74.70 & 74.40\\
      ~ & 2 & 76.80 & 77.40 & 76.20\\
      ~ & 3 & 77.50 & {\bf 78.60} & 77.70\\
      ~ & 4 & 77.60 & 78.20 & 78.10\\
      ~ & 5 & 77.50 & {\bf 78.60} & 77.90\\
      \hline
    \end{tabular}}
    \captionof{table}{\small Graph classification accuracies (\%) with various scales and feature-crossing layers on IMDB-B.}
    \label{table:scale-layer}
  \end{minipage}
  \hfill
  \begin{minipage}[b]{0.49\textwidth}
    \centering
    \includegraphics[width=1\textwidth]{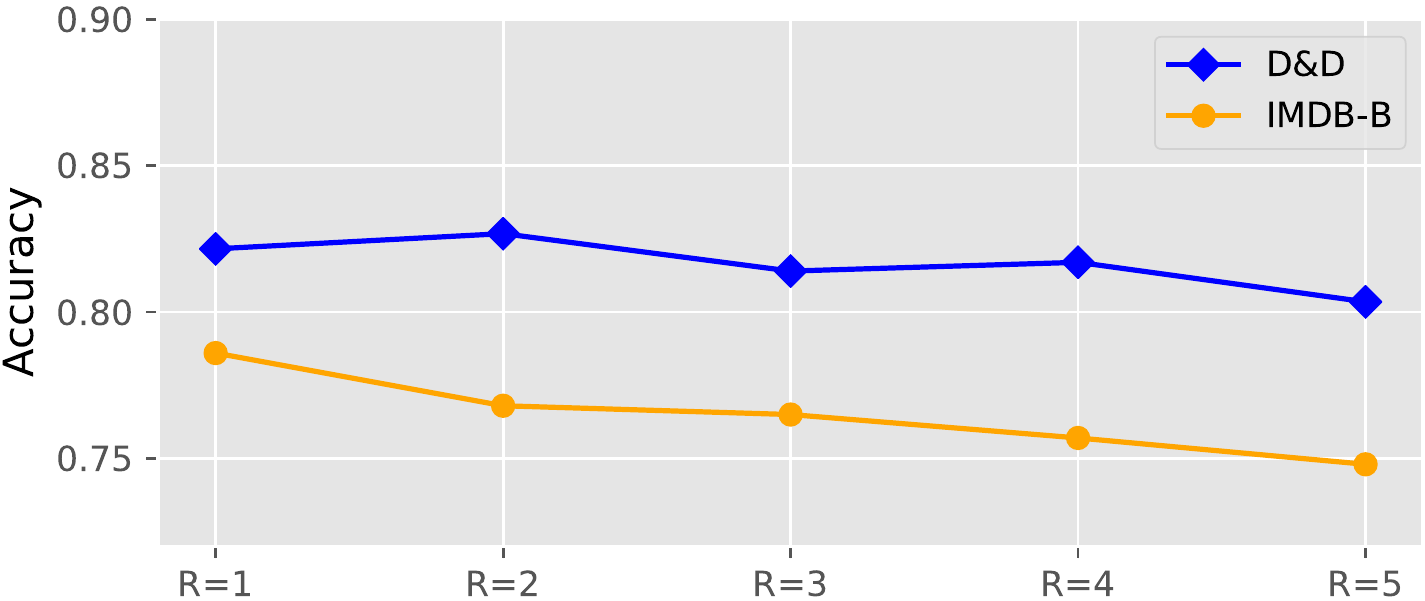}
    \caption{\small Graph classification accuracies (\%) with neighbor-hops $R$ from 1 to 5 on D\&D and IMDB-B.}
    \label{fig:Hop}
  \end{minipage}
  \vspace{-10pt}
\end{figure}

\textbf{GXN architectures.} We test the architecture with various graph scales and feature-crossing layers. Base on the dataset of IMDB-B for graph classification, we vary the number of graph scales from $1$ to $5$ and the number of feature-crossing layers from $1$ to $3$. We present the vertex classification results in Table~\ref{table:scale-layer}. We see that the architecture with $3$ graph scales and $2$ feature-crossing layers leads to the best performance. Compared to use only $1$ graph scale, using $2$ graph scales significantly improve the graph classification accuracy by $2.53\%$ on average, indicating the importance of  multiscale representations. When we use more than 3 scales, the classification results tend to be stable, indicating the redundant scales. To keep the model efficiency  and effectiveness, we adopt 3 scales of graphs. As for the number of feature-crossing layers, only using $1$ feature-crossing layer do not provide sufficient information for graph classification; while using more than 2 feature-crossing layers tends to damage model performance due to the higher model complexity.

\textbf{Hops of neighborhood in VIPool.}
To validate the effects of different ranges of neighborhood information in VIPool, we vary the neighbor-hops $R$ in $\mathcal{P}_w(\cdot)$ from $1$ to $5$ and perform graph classification on D\&D and IMDB-B. When $R$ increases, we push a vertex to represent a bigger neighborhood with more global information.
Figure~\ref{fig:Hop} shows the graph classification accuracies with various $R$ on the two datasets. We see that, for D\&D, which include graphs with relatively larger sizes (see Table~\ref{tab:graph_classification}, line 3), when $R=2$, the model achieves the best performance, reflecting that vertices could express their neighborhoods within $R=2$; while for IMDB-B with smaller graphs, vertices tend to express their 1-hop neighbors better. This reflects that VIPool achieves a flexible trade-off between local and global information through varying $R$ to adapt to various graphs. 

\begin{figure}[!t]
  \vspace{-5pt}
  \begin{center}
    \begin{tabular}{cccc}
     \includegraphics[width=0.3\columnwidth]{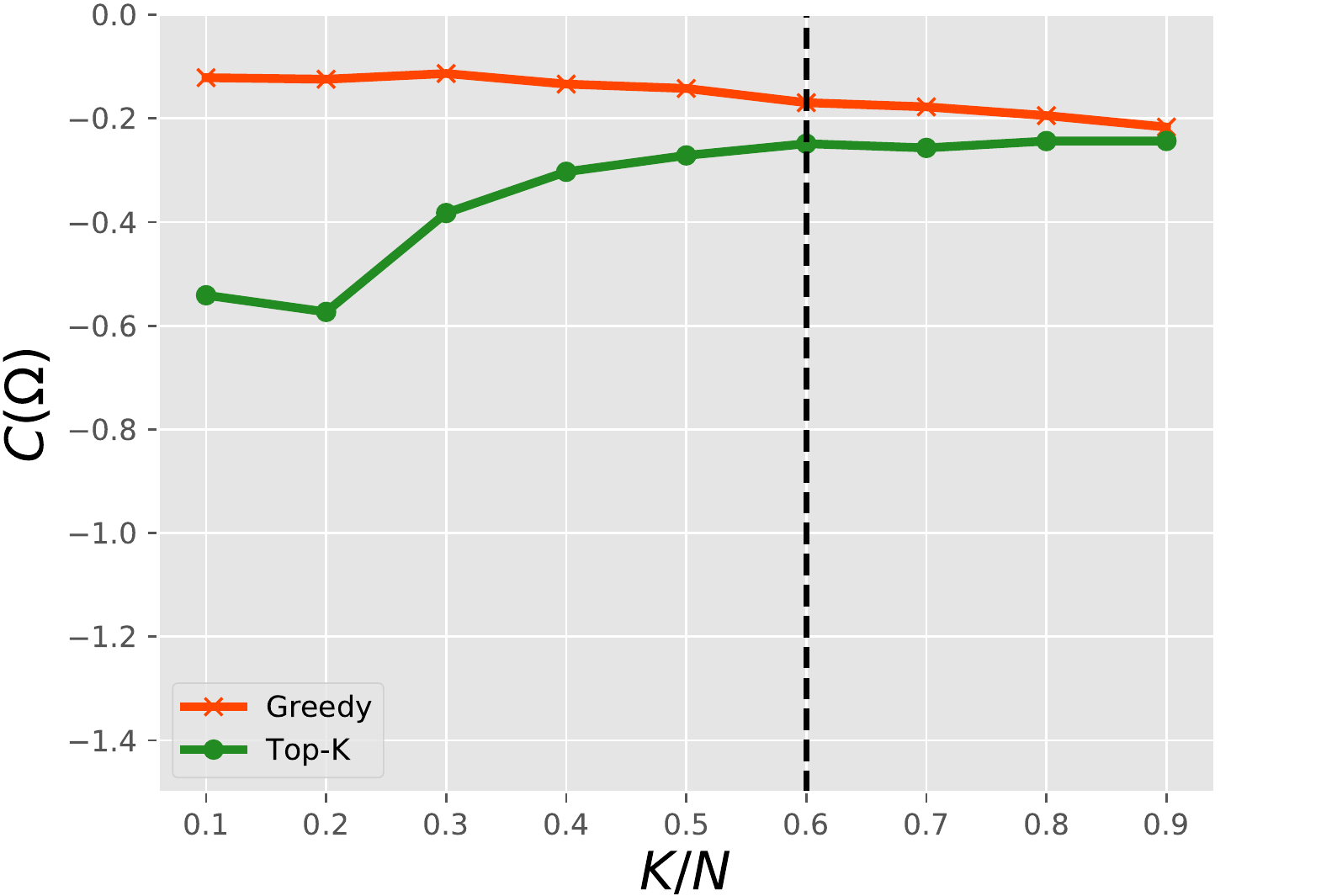}   
%   & \includegraphics[width=0.225\columnwidth]{figures/Qcurve_protein.pdf} 
   & \includegraphics[width=0.3\columnwidth]{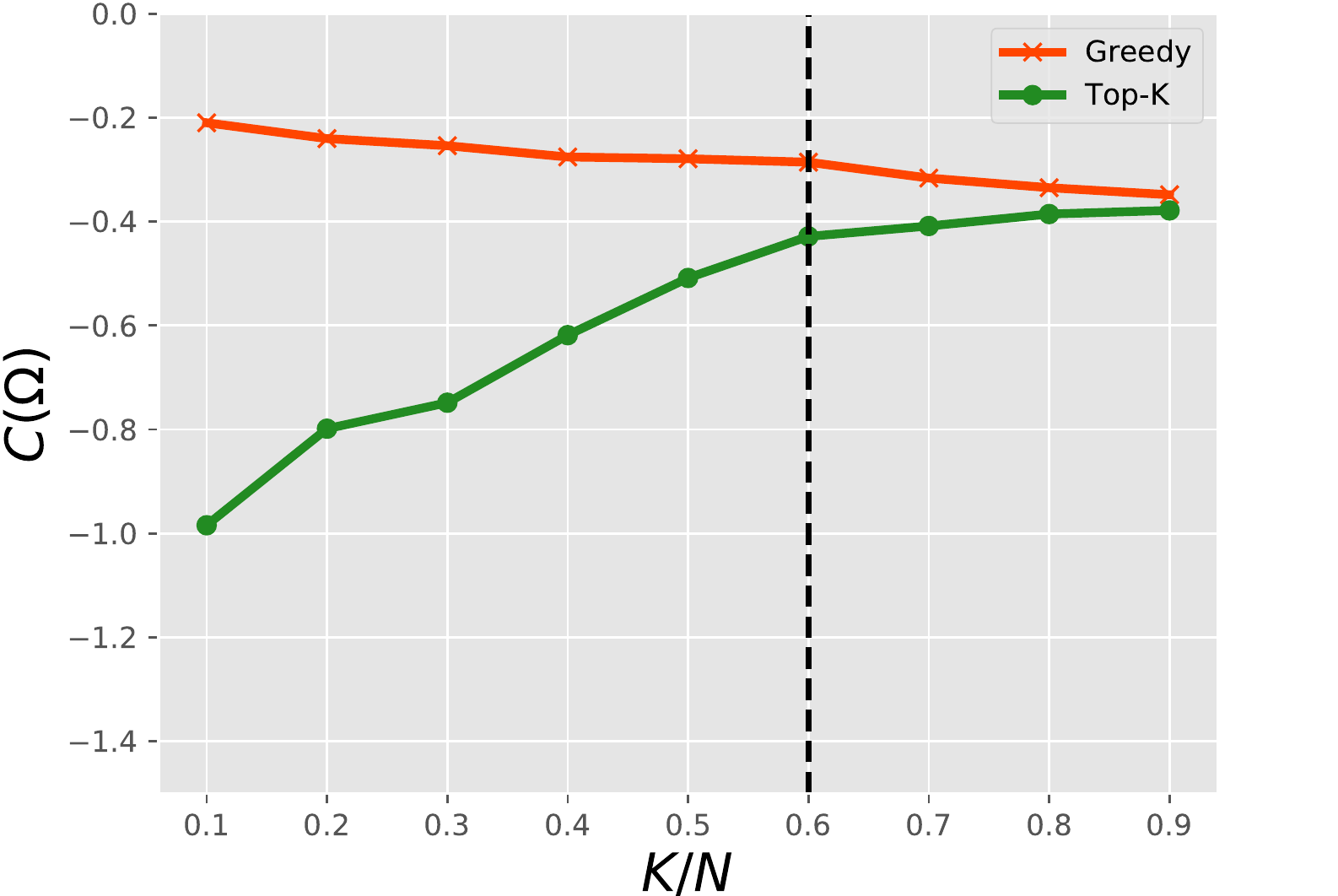}
   & \includegraphics[width=0.3\columnwidth]{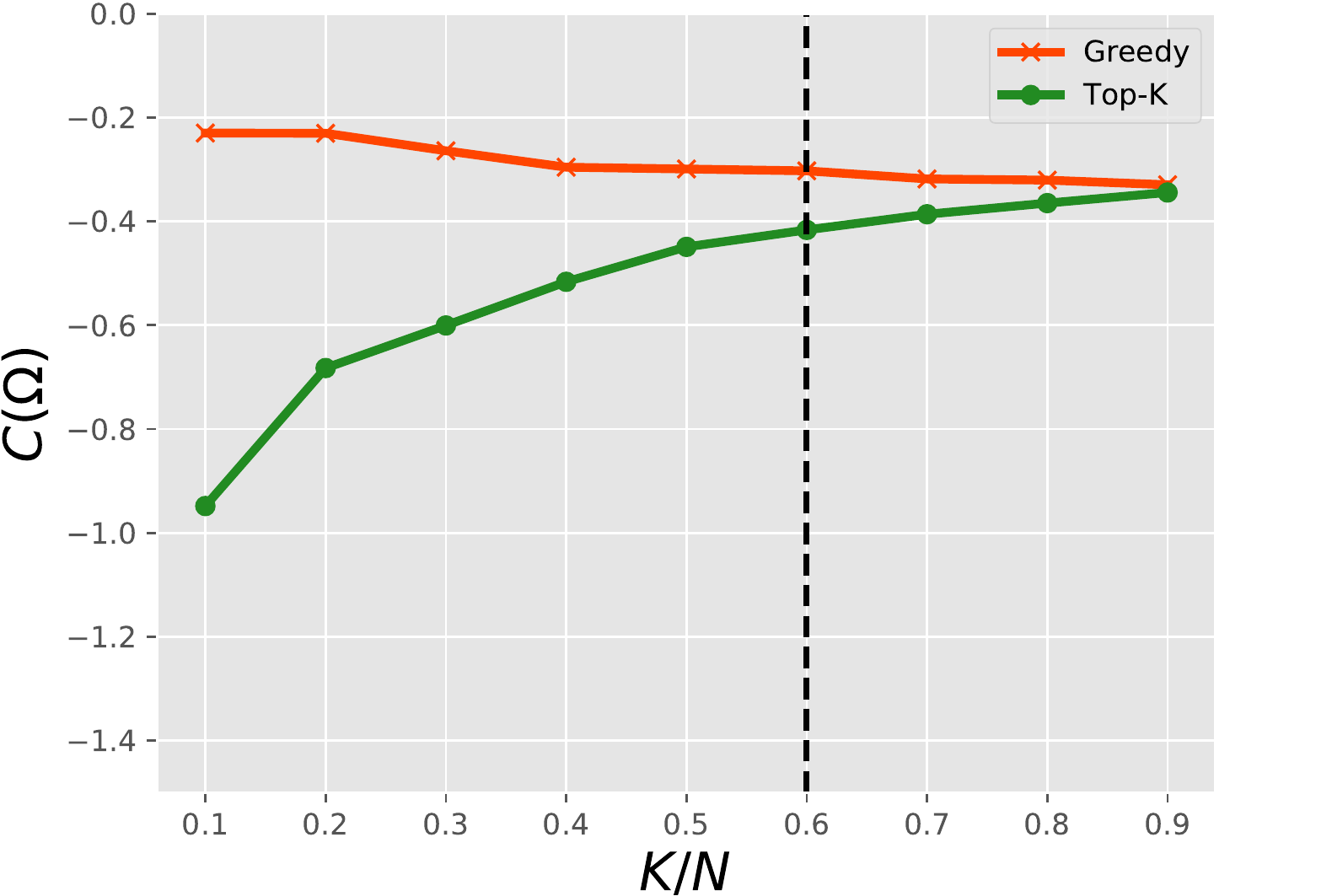}
    \\
    {\small (a) PROTEINS (bioinformatic).} &  
    {\small (b) COLLAB (social).} &
    {\small (c) Cora (citation).} 
  \end{tabular}
\end{center}
\vspace{-10pt}
\caption{\small Comparison of C values on different types of graph datasets.}
\vspace{-2pt}
\label{fig:Qcompare}
\end{figure}

\textbf{Approximation of C function.} In VIPool, to optimize problem~\eqref{eq:problem} more efficiently, we substitute the original $C(\Omega)$ by only preserving the positive term $C_{+}(\Omega)=\sum_{v\in\Omega}\log\sigma(\mathcal{S}_w({\bf h}_v, {\bf h}_{\mathcal{N}_v}))$ and maximize $C_{+}(\Omega)$ by selecting `Top-K', which also obtains the optimal solution. To see the performance gap between the original and the accelerated versions, we compare the exact value of $C(\Omega)$ with the selected $\Omega$ by optimizing $C(\Omega)$ with greedy algorithm and optimizing $C_{+}(\Omega)$ with `Top-K' method, respectively, on different types of datasets: bioinformatic, social and citation networks. We vary the ratio of the vertex selection among the global graph from $0.1$ to $0.9$. Figure~\ref{fig:Qcompare} compares the $C(\Omega)$ as a function of selection ratio with two algorithms on $3$ datasets, and the vertical dash lines denotes the boundaries where the value gaps equal to $10\%$.
We see that, when we select small percentages (e.g. $<60\%$) of vertices, the $C$ value obtained by the greedy algorithm is much higher than `Top-K' method; when we select more vertices, there are very small gaps between the two optimization algorithms, indicating two similar solutions of vertex selection. In GXN, we set the selection ratio above $60\%$ in each graph pooling. More results about the model performances varying with ratios of vertices selection are presented in Appendix.

% \begin{wraptable}{r}{5.3cm}
%     \small
% 	\centering
% 	\caption{Vertex classification accuracies and running time of different adjcacency pooling operations on Cora.}
% 	\begin{tabular}{c|ccccccc}
% 	\hline
% 	Adjacency pool & accuracy & time \\
% 	\hline
% 	Edge remove \\
% 	Kron reduction \\
% 	Clus-connect \\
% 	\hline
% 	\end{tabular}
% \end{wraptable}

\mypar{Graph structure pooling} In VIPool, we consider three implementations for graph structure pooling ${\rm Pool}_A(\cdot)$: edge-removal, Kron reduction and cluster-connection. We test these three operations for semi-supervised vertex classification on Cora and graph classification on IMDB-B, and we show the classification accuracies and time costs of the three graph structure pooling operations (denoted as `Edge-Remove', `Kron-Reduce' and `Clus-Connect', respectively) in Tables~\ref{table:poolAcora} and~\ref{table:poolAIMDB}.
\begin{figure}[!t]
\vspace{-5pt}
  \begin{minipage}[b]{0.48\textwidth}
    \small
    \centering
    \captionof{table}{\small  Vertex classification accuracies and time costs per epoch with different ${\rm Pool}_A(\cdot)$ on Cora.}
    \begin{tabular}{c|cc}
      \hline
      ${\rm Pool}_A(\cdot)$ & Accuracy (\%) & Time (s)\\
      \hline
      Edge-Remove & 84.6  & 0.44  \\
      Kron-Reduce & 85.3  & 8.36  \\
      Clus-Connect & 85.1 & 1.28  \\
      \hline
    \end{tabular}
    \label{table:poolAcora}
  \end{minipage}
  \hfill
\begin{minipage}[b]{0.48\textwidth}
    \small
    \centering
    \captionof{table}{\small Graph classification accuracies and time costs per epoch with different ${\rm Pool}_A(\cdot)$ on IMDB-B.}
    \begin{tabular}{c|cc}
      \hline
      ${\rm Pool}_A(\cdot)$ & Accuracy & Time (s)\\
      \hline
      Edge-Remove & 78.20 & 1.97  \\
      Kron-Reduce & 78.50 & 13.44  \\
      Clus-Connect & 78.60 & 3.75\\
      \hline
    \end{tabular}
    \label{table:poolAIMDB}
  \end{minipage}
  \vspace{-10pt}
\end{figure}
We see that Kron reduction or cluster-connection tend to provide the best accuracies on different datasets, but Kron reduction is significantly more expensive than the other two methods due to the  matrix inversion. On the other hand, cluster-connection provides a better tradeoff between effectiveness and efficiency and we thus consider cluster-connection as our default choice. 
\vspace{-1mm}
\section{Conclusions}
\vspace{-2mm}
This paper proposes a novel model~\emph{graph cross network} (GXN), where we construct parallel networks for feature learning at multiple scales of a graph and design novel feature-crossing layers to fuse intermediate features across multiple scales. To downsample graphs into various scales, we propose vertex infomax pooling (VIPool), selecting those vertices that maximally describe their neighborhood information.
Based on the selected vertices, we coarsen graph structures and the corresponding graph data. VIPool is optimized based on the neural estimation of the mutual information between vertices and neighborhoods. Extensive experiments show that (i) GXN outperforms most state-of-the-art methods on graph classification and vertex classification; (ii) VIPool outperforms the other pooling methods; and (iii) more intermediate fusion across scales leads to better performances.

\section*{Acknowledgement}
This work is supported by the National Key Research and Development Program of China (No. 2019YFB1804304), SHEITC (No. 2018-
RGZN-02046), 111 plan (No. BP0719010), and STCSM (No. 18DZ2270700), and State Key Laboratory of UHD Video and Audio Production and Presentation. Prof. Ivor Tsang is supported by ARC DP180100106 and DP200101328.

\section*{Broader Impact of Our Work}
In this work, we aim to propose a method for multiscale feature learning on graphs, achieving two basic but challenging tasks: graph classification and vertex classification. This work has the following potential impacts to the society and the research community.

This work could be effectively used in many practical and important scenarios such as drug molecular analysis, social network mining, biometrics, human action recognition and motion prediction, etc., making our daily life more convenient and efficient. Due to the ubiquitous graph data, in most cases, we can try to construct multiscale graphs to comprehensively obtain rich detailed, abstract, and even global feature representations, and effectively improve downstream tasks. 

Our network structure can not only solve problem of feature learning with multiple graph scales, but also can be applied to the pattern learning of heterogeneous graphs, or other cross-modal or cross-view machine learning scenarios. This is of great significance for improving the ability of pattern recognition, feature transfer, and knowledge distillation to improve the computational efficiency.

At the same time, this work may have some negative consequences. For example, in social networks, it is uncomfortable even dangerous to use the models based on this work to over-mine the behavior of users, because the user's personal privacy and information security are crucial; companies should avoid mining too much users' personal information when building social platforms, keeping a safe internet environment.

{\small
\bibliographystyle{plainnat}
\bibliography{egbib}
}

\newpage

\section*{Appendix A. Mutual Information Neural Estimation for Vertex Selection}
An individual vertex is fully identified through its feature, which works as the vertex attribute. Given an vertex set $\mathcal{V}$ that contains all the vertices on the graph and a vertex subset $\Omega\subset\mathcal{V}$ which contains the selected vertices, we let a random variable ${\bf v}$ to represent the vertex feature when we randomly pick a vertex from $\Omega$. Then we define the probability distribution of ${\bf v}$ as
\begin{equation*}
    P_{\bf v} ~=~ P({\bf v}={\bf x}_v), \ \ \ \ \forall v\in\Omega
\end{equation*}
where ${\bf x}_v$ is the feature value when we pick vertex $v$.

The neighborhood of any vertex $u\in\Omega$ is defined as $\mathcal{N}_u$,  which is the subgraph containing the vertices in $\mathcal{V}$ whose geodesic distances to the central vertex $u$ are no greater than a threshold $R$,
 i.e. $\mathcal{N}_v = G(\{ u \}_{d(u, v)\le R}, \mathbf{A}_{\{u\}, \{u\}})$. Let a random variable ${\bf n}$ be the neighborhood features when we randomly pick a central vertex from $\Omega$, then we define the probability distribution of ${\bf n}$ as
\begin{equation*}
    P_{\bf n} ~=~ P({\bf n}={\bf y}_{\mathcal{N}_u}), \ \ \ \ \forall u \in \Omega
\end{equation*}
where ${\bf y}_{\mathcal{N}_u}=[{\bf A}_{\mathcal{N}_u, \mathcal{N}_u}, \{{\bf x}_{\nu}\}_{\nu\in\mathcal{N}_u}]$ denotes the neighborhood feature value when we pick vertex $u$'s neighborhood, including both the internal connectivity information and contained vertex features in the neighborhood $\mathcal{N}_u$.

Therefore, we define the joint distribution of the random variables of vertex features and neighborhood features, which is formulated as
\begin{equation*}
    P_{{\bf v},{\bf n}} ~=~ P({\bf v}={\bf x}_v, {\bf n}={\bf y}_{\mathcal{N}_v}), \ \ \ \ \forall v \in \Omega
\end{equation*}
where the joint distribution reflects probability that we randomly pick the corresponding vertex feature and neighborhood feature of the same vertex $v$ together.

The \emph{mutual information} between the vertex features and the neighborhood features is defined as the KL-divergence between the joint distribution $P_{{\bf v},{\bf n}}$ and the product of the marginal distributions of two random variable, $P_{\bf v}\otimes P_{\bf n}$; that is
\begin{eqnarray*}
    I^{(\Omega)} \left( {\bf v}, {\bf n} \right)
    & = & D_{\rm KL} \left( P_{{\bf v},{\bf n}} || P_{\bf v}\otimes P_{\bf n} \right),
\end{eqnarray*}
This mutual information measures of the mutual dependency between vertices and neighborhoods in the selected vertex subset $\Omega$. The KL divergence admits the $f$-representation~\cite{pmlr-v80-belghazi18a},
\begin{eqnarray}
\label{eq:f-KL}
D_{\rm KL} 
\left( P_{{\bf v},{\bf n}} || P_{\bf v}\otimes P_{\bf n} \right) \ \geq \
\sup_{T\in\mathcal{T}} \left\{  
    \mathbb{E}_{{\bf x}_v, {\bf y}_{\mathcal{N}_v}  \sim  P_{{\bf v}, {\bf n}}} 
    \left [ T( \mathbf{x}_v, \mathbf{y}_{\mathcal{N}_v} ) \right ] 
    - \mathbb{E}_{ {\bf x}_v \sim  P_{{\bf v}}, {\bf y}_{\mathcal{N}_u} \sim  P_{\bf n} } 
    \left [ e^{T \left( \mathbf{x}_v, \mathbf{y}_{\mathcal{N}_u}\right)-1} \right ] 
    \right\},
\end{eqnarray}
where $\mathcal{T}$ is an arbitrary class of functions that maps a pair of vertex features and neighborhood features to a real value, and here we use $T(\cdot,\cdot)$ to compute the dependency of two features. It could be a tight lower-bound of mutual information if we search any possible function $T\in\mathcal{T}$.

{Note that the main target here is to propose a vertex-selection criterion based on quantifying the dependency between vertices and neighborhood. Therefore instead of computing the exact mutual information based on KL divergence, we can use non-KL divergences to achieve favourable flexibility and convenience in optimization. Both non-KL and KL divergences can be formulated based on the same $f$-representation framework. Here we start from the general $f$-divergence between the joint distribution and the product of marginal distributions of vertices and neighborhoods.}
\begin{eqnarray*}
D_{f} \left( P_{{\bf v},{\bf n}} || P_{\bf v} \otimes P_{\bf n} \right)  \ = \ \int P_{{\bf v}}P_{{\bf n}}
           f \left( \frac{P_{{\bf v},{\bf n}} }
                         {P_{\bf v} P_{\bf n} } \right) 
                         d {\bf x}_v d {\bf y}_{\mathcal{N}_v}
\end{eqnarray*}
where $f(\cdot)$ is a convex and lower-semicontinuous divergence function. when $f(x)=x \log x$, the $f$-divergence is specificed as KL divergence. The function $f(\cdot)$ has a convex conjugate function $f^*(\cdot)$, i.e. 
$f^*(t)=\sup_{x\in{\rm dom}_f}\{xt-f(x)\}$, where ${\rm dom}_f$ is the definition domain of $f(\cdot)$. 
Note that the two functions $f(\cdot)$ and $f^*(\cdot)$ is dual to each other.
According to the Fenchel conjugate~{\cite{convexanalysis}}, the $f$-divergence can be modified as
\begin{eqnarray*}
    D_f(P_{{\bf v},{\bf n}} || P_{\bf v} \otimes P_{\bf n}) 
    & = & 
    \int P_{\bf x}P_{\bf n}
    \sup_{t\in {\rm dom}_{f^*} } 
    \left \{ t \frac{P_{{\bf x},{\bf n}}}{P_{\bf v} P_{\bf n}} -
    f^*(t) \right \} \\
    & \geq & 
    \sup_{T\in\mathcal{T}} 
    \left \{
    \mathbb{E}_{P_{{\bf v},{\bf n}}} 
    [T({\bf x}_{v}, {\bf y}_{{\mathcal{N}_v}})] -
    \mathbb{E}_{P_{{\bf v}},P_{{\bf n}}} 
    [f^*(T({\bf x}_{v}, {\bf y}_{{\mathcal{N}_u}}))]
    \right \}
\end{eqnarray*}
where $\mathcal{T}$ denotes any functions that map vertex and neighborhood features to a scalar, and the function $T(\cdot,\cdot)$ works as a variational representation of $t$. We further use an activation function $a:\mathbb{R}\to{\rm dom}_{f^*}$ to constrain the function value; that is $T(\cdot,\cdot)\to a(T(\cdot,\cdot))$. Therefore, we have
\begin{eqnarray*}
    D_f(P_{{\bf v},{\bf n}} || P_{\bf v} \otimes P_{\bf n}) 
    & \geq & 
    \sup_{T\in\mathcal{T}} 
    \left \{
    \mathbb{E}_{P_{{\bf v},{\bf n}}} 
    [a(T({\bf x}_{v}, {\bf y}_{{\mathcal{N}_v}}))] -
    \mathbb{E}_{P_{{\bf v}},P_{{\bf n}}} 
    [f^*(a(T({\bf x}_{v}, {\bf y}_{{\mathcal{N}_u}})))]
    \right \}
\end{eqnarray*}
since the  $a(T(\cdot,\cdot))$ is also in $\mathcal{T}$ and its value is in ${\rm dom}_{f^*}$, the optimal solution satisfies the equation.
{Suppose that the divergence function is $f(x)=x \log x$,  the conjugate divergence function is $f^*(t)= \exp(t-1)$ and the activation function is $a(x)=x$, we can obtain the $f$-representation of KL divergence; see Eq.~\eqref{eq:f-KL}.} Note that the form of activation function is not unique, and we aim to find a proper one that helps to derivation and computation.

Here, we consider another form of divergence based on $f$-representation; that is, GAN-like divergence, where we have specific form of divergence function $f(x) = x \log x - (x+1) \log (x+1)$ and conjugate divergence function $f^*(t) = -\log (1-\exp(t))$~{\cite{NIPS2016_6066}}. We let the activation be $a(\cdot)=-\log(1+\exp(\cdot))$.  The GAN-like divergence is formulated as
\begin{eqnarray*}
&& D_{\rm GAN} 
\left( P_{{\bf v},{\bf n}} 
|| P_{\bf v} \otimes P_{\bf n} \right) \\
& \geq &
\sup_{T\in\mathcal{T}} 
    \left \{
    \mathbb{E}_{P_{{\bf v},{\bf n}}} 
    [a(T({\bf x}_{v}, {\bf y}_{{\mathcal{N}_v}}))] -
    \mathbb{E}_{P_{{\bf v}},P_{{\bf n}}} 
    [f^*(a(T({\bf x}_{v}, {\bf y}_{{\mathcal{N}_u}})))]
    \right \} \\
& = &
\sup_{T\in\mathcal{T}}
    \left \{
    \mathbb{E}_{P_{{\bf v},{\bf n}}} 
    [-\log(1+\exp(-T({\bf x}_v, {y}_{\mathcal{N}_v})))]+
    \mathbb{E}_{P_{{\bf v}},P_{{\bf n}}}
    \log(1-\exp(-\log(1+e^{T({\bf x}_v,{\bf y}_{\mathcal{N}_u})})))
    \right \} \\
& = &
\sup_{T\in\mathcal{T}}
    \left \{
    \mathbb{E}_{P_{{\bf v},{\bf n}}}
    \log
    \frac{1}{1+e^{-T({\bf x}_v, {y}_{\mathcal{N}_v})}} +
    \mathbb{E}_{P_{{\bf v}},P_{{\bf n}}}
    \log(1-\frac{1}{1+e^{-T({\bf x}_v,{\bf y}_{\mathcal{N}_u})}})
    \right \} \\
& = &
\sup_{T\in\mathcal{T}} \bigg\{  
    \mathbb{E}_{P_{{\bf v}, {\bf n}}} 
    \big[ \log \sigma \left( T( \mathbf{x}_v, \mathbf{y}_{\mathcal{N}_v} ) \right) \big] +
    \mathbb{E}_{P_{{\bf v}},P_{{\bf n}}}
    \big[ \log \left( 1 - \sigma \left( T \left( \mathbf{x}_v, \mathbf{y}_{\mathcal{N}_u}\right) \right) \right) \big] 
    \bigg\}
% \\
% & \approx &
% \max_{w} \bigg\{  
%     \mathbb{E}_{v, \mathcal{N}_v  \sim  P_{\mathcal{V}, \mathcal{N}}} 
%     \big[ \log \sigma \left( T_w( \mathbf{x}_v, \mathbf{y}_{\mathcal{N}_v} ) \right) \big] 
%     + \mathbb{E}_{ v \sim  P_{\mathcal{V}}, \mathcal{N}_u \sim  P_{\mathcal{N}} } \big[ \log \left( 1 - \sigma \left( T_w \left( \mathbf{x}_v, \mathbf{y}_{\mathcal{N}_u}\right) \right) \right) \big] 
%     \bigg\}
% \\
% & = &
% \max_{w} \bigg\{   \frac{1}{|\mathcal{V}|} \sum_{v \in \mathcal{V}}  \log \sigma \left( \mathcal{S}_w( {\bf h}_v, {\bf h}_{\mathcal{N}_v} ) \right)
%   + \frac{1}{|\mathcal{V}|^2} \sum_{v,u \in  \mathcal{V} }   \log  \big( 1 - \sigma \left( \mathcal{S}_w( {\bf h}_v, {\bf h}_{\mathcal{N}_u} ) \right)  \big)     \bigg\}.
\end{eqnarray*}
where $\sigma(\cdot)$ is the sigmoid function that maps a real value into the range of $(0,1)$.
Eventually, the GAN-like divergence converts the $f$-divergence to a binary cross entropy, which is similar to the objective function to train the discriminator in GAN~{\cite{NIPS2014_5423}}.

To determine the form of the function $T(\cdot,\cdot)$, we parameterized  $T(\cdot,\cdot)$ by trainable neural networks rather than design it manually. The parameterized function is denoted as $T_w(\cdot,\cdot)$, where $w$ generally denotes the parameterization. In this work, $T_w(\cdot,\cdot)$ is constructed with three trainable functions: 1) A vertex embedding function $\mathcal{E}_w(\cdot)$; 2) A neighborhood embedding function $P_w(\cdot)$; and 3) a vertex-neighborhood affinity function $C_w(\cdot, \cdot)$; which are formulated as
\begin{eqnarray*}
T_w ( \mathbf{x}_v, \mathbf{y}_{\mathcal{N}_{u}} ) 
& = & \mathcal{S}_w(
                    \mathcal{E}_w({\bf x}_v), \mathcal{P}_w({\bf y}_{\mathcal{N}_{u}})) \\
& = & \mathcal{S}_w \Bigg( \mathcal{E}_w(\mathbf{x}_v), \frac{1}{R}\sum_{r=0}^{R}\sum_{\nu\in\mathcal{N}_{u}}
\left((\widetilde{\bf D}^{-1/2}
      \widetilde{\bf A}
      \widetilde{\bf D}^{-1/2})^r\right)_{\nu,u}
{\bf W}^{(r)}
\mathcal{E}_w(\mathbf{x}_{\nu})
\Bigg).
\end{eqnarray*}
where $\mathcal{E}_w(\cdot)$ is modeled by a Multi-layer perceptron (MLP), $\mathcal{P}_w(\cdot)$ is modeled by a $R$-hop graph convolution layer and $\mathcal{S}_w(\cdot, \cdot)$ is also modeled by an MLP. In $\mathcal{P}_w(\cdot)$,
$\widetilde{\bf A}={\bf A}+{\bf I}$ is  the self-connected graph adjacency matrix and $\widetilde{\bf D}$ is the degree matrix of $\widetilde{\bf A}$; ${\bf W}^{(r)}\in\mathbb{R}^{d \times d}$ is the trainable weight matrix associated with the $r$th hop of neighborhood. The neighborhood embedding function $\mathcal{P}_w(\cdot)$ aggregates neighborhood information with in a geodesic distance threshold $R$. Note that $\mathcal{P}_w(\cdot)$ separately use neighborhood features ${\bf y}_{\mathcal{N}_u}$ in form of connectivity information and vertex features.

In this way, the GAN-like-divergence-based mutual information between graph vertices and neighborhoods can be represented with the parameterized GAN-like divergence, which is a variational divergence and works as a lower bound of of the theorical GAN-like-divergence-based mutual information; that is,
\begin{eqnarray*}
    % I^{(\Omega)}({\bf v},{\bf n}) 
    % & \geq & 
    I^{(\Omega)}_{\rm GAN}({\bf v},{\bf n}) 
    &=& D_{\rm GAN}(P_{{\bf v},{\bf n}}||P_{{\bf v}}\otimes P_{{\bf n}}) 
    ~\geq~  \widehat{I}^{(\Omega)}_{\rm GAN}({\bf v},{\bf n}) \\
    & = & 
    \max_{w} \bigg\{  
    \mathbb{E}_{P_{{\bf v}, {\bf n}}} 
    \big[ \log \sigma \left( T_w( \mathbf{x}_v, \mathbf{y}_{\mathcal{N}_v} ) \right) \big] +
    \mathbb{E}_{P_{{\bf v}},P_{{\bf n}}}
    \big[ \log \left( 1 - \sigma \left( T_w \left( \mathbf{x}_v, \mathbf{y}_{\mathcal{N}_u}\right) \right) \right) \big] 
    \bigg\} \\
    & = &
    \max_{w}
    \frac{1}{|\Omega|}
    \sum_{v\in\Omega}\log\sigma
    (\mathcal{T}_w({\bf x}_v,{\bf y}_{\mathcal{N}_v})) +
    \frac{1}{|\Omega|^2}
    \sum_{(v,u)\in\Omega}\log
    (1-\sigma(\mathcal{T}_w({\bf x}_v,{\bf y}_{\mathcal{N}_u})))
\end{eqnarray*}
{Since we consider the dependency between vertices and neighborhoods within a specific vertex set, the possible outcomes
of the joint distribution and the two marginal distributions are countable. We thus use the summation to aggregate all the possible cases.}
To maximize $\widehat{I}^{(\Omega)}_{\rm GAN}({\bf v},{\bf n})$ by training the internal function in $T_w(\cdot,\cdot)$, that is, $\mathcal{E}_w(\cdot)$, $\mathcal{P}_w(\cdot)$, and $\mathcal{S}_w(\cdot,\cdot)$, we can maximally approximate the mutual information between individual vertex and neighborhood for vertex selection in our VIPool. Note that the value of $\widehat{I}^{(\Omega)}_{\rm GAN}({\bf v},{\bf n})$ is not very close to the exact KL-divergence-based mutual information, but it has the consistency to $I^{(\Omega)}({\bf v},{\bf n})$ to 
reflect the pair of vertex-neighborhood with high or low mutual information, leading to effective vertex selection.

\section*{Appendix B. Detailed Information of Experimental Graph Datasets}
Here we show more details about the  graph datasets used in our experiments of both graph classification and vertex classification. We first show the six datasets for graph classification in Table~\ref{tab:graphclassificationdata}.
\begin{table}[t]
    \centering
    \footnotesize
    \caption{\small The detailed information of graph datasets used in the experiments of graph classification}
    \begin{tabular}{c|cccccc}
        \specialrule{0.08em}{1pt}{1pt}
        Dataset & IMDB-B & IMDB-M & COLLAB & D\&D & PROTEINS  & ENZYMES \\
        \hline
        \# Graphs & 1000 & 1500 & 5000 & 1178 & 1113 & 600 \\
        \# Classes & 2 & 3 & 3 & 2 & 2 & 6 \\
        Max \# Vertices & 139 & 89 & 492 & 5748 & 620 & 126 \\
        Min \# Vertices & 12 & 7 & 32 & 30 & 4 & 2 \\
        Avg. \# Vertices & 19.77 & 13.00 & 74.49 & 284.32 & 39.06 & 32.63 \\
        \# Train Graphs & 900 & 1350 & 4501 & 1061 & 1002 & 540 \\
        \# Test Graphs & 100 & 150 & 499 & 111 117 & & 60 \\
        Vertex Dimensions & 1 & 1 & 1 & 82 & 3 & 3 \\
        Max Degrees & 66 & 60 & 370 & - & - & - \\
        \specialrule{0.08em}{1pt}{1pt}
    \end{tabular}
    \label{tab:graphclassificationdata}
\end{table}
We see that, we show the numbers of graphs, graph classes, vertices, numbers of graphs in training/test datasets and feature dimensions of all the six datasets. Note that, three social network datasets, IMDB-B, IMDB-M and COLLAB do not provide specific vertex features, where the vertex dimension is denoted as 1 and the maximum vertex degrees are shown in addition. In our experiments, we use one-hot vectors to encode the vertex degrees in these three datasets as their vertex features which explicitly contains the structure information.

We then show the details of three citation network datasets used in the experiments of vertex classification in Table~\ref{tab:vertexclassificationdata}.
\begin{table}[t]
    \centering
    \footnotesize
    \caption{\small The detailed information of graph datasets used in the experiments of vertex classification}
    \setlength{\tabcolsep}{8.60mm}{
    \begin{tabular}{c|ccc}
        \specialrule{0.08em}{1pt}{1pt}
        Dataset & Cora & Citeseer & Pubmed \\
        \hline
        \# Vertices & 2708 & 3327 & 19717 \\ 
        \# Edges & 5429 & 4732 & 44338 \\ 
        \# Classes & 7 & 6 & 3 \\ 
        Vertex Dimension & 1433 & 3703 & 500 \\ 
        \# Train Vertices (full-sup.) & 1208 & 1827 & 18217 \\
        \# Train Vertices (semi-sup.) & 140 & 120 & 60 \\
        \# Valid. Vertices & 500 & 500 & 500 \\
        \# Test Vertices & 1000 & 1000 & 1000 \\
        \specialrule{0.08em}{1pt}{1pt}
    \end{tabular}}
    \label{tab:vertexclassificationdata}
\end{table}
We see that, we present the numbers of vertices, edges, vertex classes and feature dimensions of the three datasets, as well as we show the separations of training/validation/test sets, where `\# Train Vertices (full-sup.)' denotes the number of training vertices for full-supervised vertex classification and `\# Train Vertices (semi-sup.)' denotes the number of training vertices for semi-supervised vertex classification.

\section*{Appendix C. More GXN Variants for Vertex Classification}
Here we show more results of vertex classification of more variants of the proposed GXN associated with different pooling methods; that is, we test different pooling methods with the same GXN model framework, where the pooling methods include gPool~\cite{ICML2019_Gao}, SAGPool~\cite{ICML2019_Lee} and AttPool~\cite{Huang_2019_ICCV}. The full-supervised and semi-supervised vertex classification accuracies of different algorithms on three citation networks are shown in Table~\ref{tab:node_classification}. 
\begin{table}[t]
  \centering
  \scriptsize
  \caption{Vertex classification accuracies (\%) of different methods, where `full-sup.' and `semi-sup.' denote the scenarios of full-supervised and semi-supervised vertex classification, respectively.}
  \setlength{\tabcolsep}{3.60mm}{
  \begin{tabular}{c|cc|cc|cc}
  \specialrule{0.08em}{0pt}{1pt}
  Dataset & \multicolumn{2}{c|}{Cora} & \multicolumn{2}{c|}{Citeseer} & \multicolumn{2}{c}{Pubmed} \\
  % \specialrule{0.05em}{1pt}{1pt}
  \# Vertices (Classes) & \multicolumn{2}{c|}{2708 (7)} & \multicolumn{2}{c|}{3327 (6)} & \multicolumn{2}{c}{19717 (3)} \\
  %\specialrule{0.05em}{1pt}{1pt}
  Supervision & full-sup. & semi-sup. & full-sup. & semi-sup. & full-sup. & semi-sup.\\

  \specialrule{0.05em}{1pt}{1pt}
  DeepWalk~\cite{DBLP:journals/corr/PerozziAS14} & 78.4 $\pm$ 1.7 & 67.2 $\pm$ 2.0 & 68.5 $\pm$ 1.8 & 43.2 $\pm$ 1.6 & 79.8 $\pm$ 1.1 & 65.3 $\pm$ 1.1 \\
  ChebNet~\cite{NIPS2016_6081} & 86.4 $\pm$ 0.5 & 81.2 $\pm$ 0.5 & 78.9 $\pm$ 0.4 & 69.8 $\pm$ 0.5 & 88.7 $\pm$ 0.3 & 74.4 $\pm$ 0.4 \\
  GCN~\cite{ICLR2017_Kipf} & 86.6 $\pm$ 0.4 & 81.5 $\pm$ 0.5 & 79.3 $\pm$ 0.5 & 70.3 $\pm$ 0.5 & 90.2 $\pm$ 0.3 & 79.0 $\pm$ 0.3\\
  GAT~\cite{velickovic2018graph} & 87.8 $\pm$ 0.7 & 83.0 $\pm$ 0.7 & 80.2 $\pm$ 0.6 & 73.5 $\pm$ 0.7 & 90.6 $\pm$ 0.4 & 79.0 $\pm$ 0.3 \\
  FastGCN~\cite{chen2018fastgcn} & 85.0 $\pm$ 0.8 & 80.8 $\pm$ 1.0 & 77.6 $\pm$ 0.8 & 69.4 $\pm$ 0.8 & 88.0 $\pm$ 0.6 & 78.5 $\pm$ 0.7 \\
  ASGCN~\cite{NIPS2018_7707} & 87.4 $\pm$ 0.3 & - & 79.6 $\pm$ 0.2 & - & 90.6 $\pm$ 0.3 & -\\
  Graph U-Net~\cite{ICML2019_Gao} & - & 84.4 & - & 73.2 & - & 79.6 \\
  \specialrule{0.05em}{0pt}{1pt}
  GXN & {\bf 88.9 $\pm$ 0.4} & {\bf 85.1 $\pm$ 0.6} & {\bf 80.9 $\pm$ 0.4} & {\bf 74.8 $\pm$ 0.4} & {\bf 91.8 $\pm$ 0.3} & {\bf 80.2 $\pm$ 0.3} \\
  GXN (gPool)   & 88.0 $\pm$ 0.4 & 84.4 $\pm$ 0.6 & 79.7 $\pm$ 0.5 & 74.4 $\pm$ 0.6 & 90.6 $\pm$ 0.4 & 79.8 $\pm$ 0.4 \\
  GXN (SAGPool) & 87.8 $\pm$ 0.6 & 84.7 $\pm$ 0.4 & 80.0 $\pm$ 0.5 & 74.2 $\pm$ 0.4 & 90.9 $\pm$ 0.3 & 80.1 $\pm$ 0.3 \\
  GXN (AttPool) & 88.4 $\pm$ 0.3 & 84.6 $\pm$ 0.5 & 80.6 $\pm$ 0.4 & 74.5 $\pm$ 0.5 & 91.3 $\pm$ 0.3 & {\bf 80.2 $\pm$ 0.4} \\
  \specialrule{0.08em}{1pt}{1pt}
  \end{tabular}}
  \label{tab:node_classification}
%   \vspace{-12pt}
\end{table}
We see that, comprared to the previous pooling methods, the proposed GXN which uses VIPool could provide higher average classification accuracies for both full-supervised and semi-supervised vertex classification. Different GXN variants with different pooling methods tend to consistently outperform most state-of-the-art models for vertex classification, reflecting the effectiveness of the proposed GXN architecture.

\section*{Appendix D. Use A Few Selected Vertices for Semi-supervised Vertex Classification Training}
Here we consider active-sample-based semi-supervised classification, where we are allowed to select a few vertices and obtain their corresponding labels as supervision to train a classifier for vertex classification. In other words, we actively select training data in a semi-supervised classification task.
Intuitively, since a graph structure is highly irregular, selecting a few informative vertices would potentially significantly improve the overall classification accuracy. Here we compare the proposed VIPool to random sampling. Note that for this task, we cannot compare with other graph pooling methods. The reason is that previous pooling pooling methods need a subsequent task to provide an explicit supervision; however, the vertex selection here should be blind to the final classification labels. The proposed VIPool is rooted in mutual information neural estimation and can be trained in either an unsupervised or supervised setting.

\begin{figure}[!t]
  \vspace{-5pt}
  \begin{center}
    \begin{tabular}{cc}
     \includegraphics[width=0.48\columnwidth]{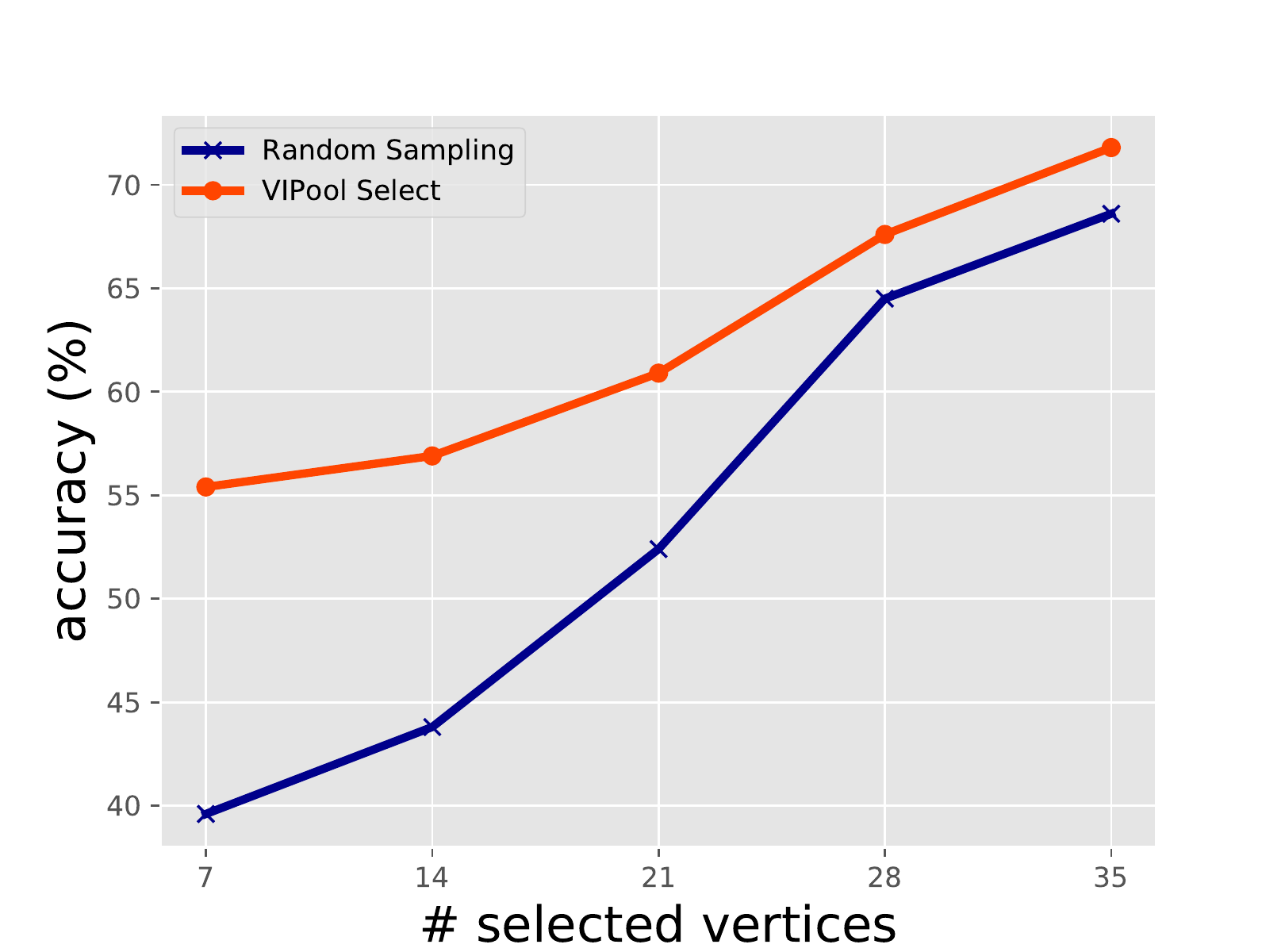}   
   & \includegraphics[width=0.48\columnwidth]{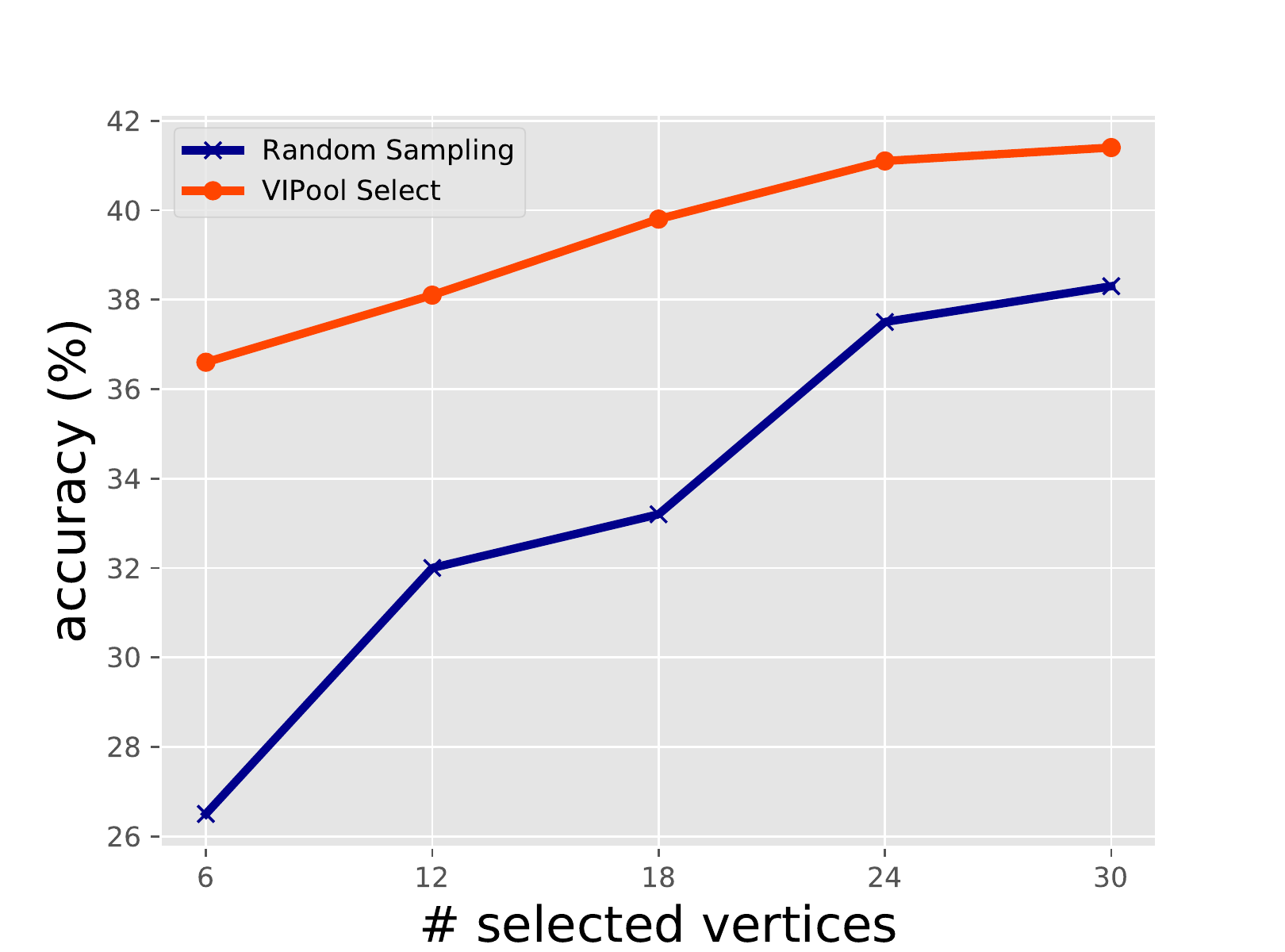}
    \\
    {\small (a) Semi-supervised vertex classification on Cora.} &  
    {\small (b) Semi-supervised vertex classification on Citeseer.}
  \end{tabular}
\end{center}
\vspace{-10pt}
\caption{\small Comparison of semi-supervised vertex classification accuracies with a few selected and labeled data by using different vertex selection methods.}
\vspace{-2pt}
\label{fig:afewvertex}
\end{figure}

Specifically, given a graph, such as a citation network, Cora or Citeseer, 
we aim to show the classification accuracy as a function of the number of selected vertices. For example, there are $7$ classes in Cora, we can select $7, 14, 21, 28$ and $35$ vertices ($1, 2, 3, 4$ and $5$ times of $7$) and use their ground-truth labels as supervision for semi-supervised vertex classification. As for Citeseer, there are $6$ classes and we can select $6, 12, 18, 24$ and $30$ vertices. During evaluation, we test the performances on all the unselected vertices. We compare two method for vertex selection and classification: 1) the proposed {\bf VIPool} method, where we use greedy algorithm to optimize $C(\Omega)$ for vertex selection; and 2) {\bf Random Sampling}, where we randomly select each vertex with the same probability on the whole graph. We conduct semi-supervised vertex classification on the datasets of Cora and Citeseer.
Figure~\ref{fig:afewvertex} shows the
 the classification accuracies varying with the numbers of selected vertices for  two vertex selection methods.
We see that, when we select only a few vertices, such as fewer than $3$ times of the number of vertex classes (i.e. $21$ for Cora and $18$ for Citeseer), the proposed VIPool method could select much more informative vertices than randomly sampling the same number of vertices, leading to over $10\%$ higher vertex classification accuracies. If we select more vertices by using the two vertex selection methods, the classification results corresponding to the two methods become closer to each other, indicating that a large number of selected vertices  tend to potentially provide sufficient information to represent the rich patterns of the graphs and we could obtain more similar classification results than only selecting a few vertices.

\section*{Appendix E. Illustration of Vertex Selection}
To show the pooling effects of different pooling algorithms, we conduct a toy experiments to reconstruct three spatial mesh graphs with an encoder-decoder model. The encoder employs different pooling methods to squeeze the original graph into a few vertices (10 vertices) and the decoder attempt to reconstruct the original graphs based on the pooled vertex features and graph structures. To train the encoder-decoder model, we use an L2-norm loss to measure the distances between the vertex coordinates of reconstructed graphs and ground-truth graphs.

\begin{figure}[t]
    \centering
    \includegraphics[width=0.8\textwidth]{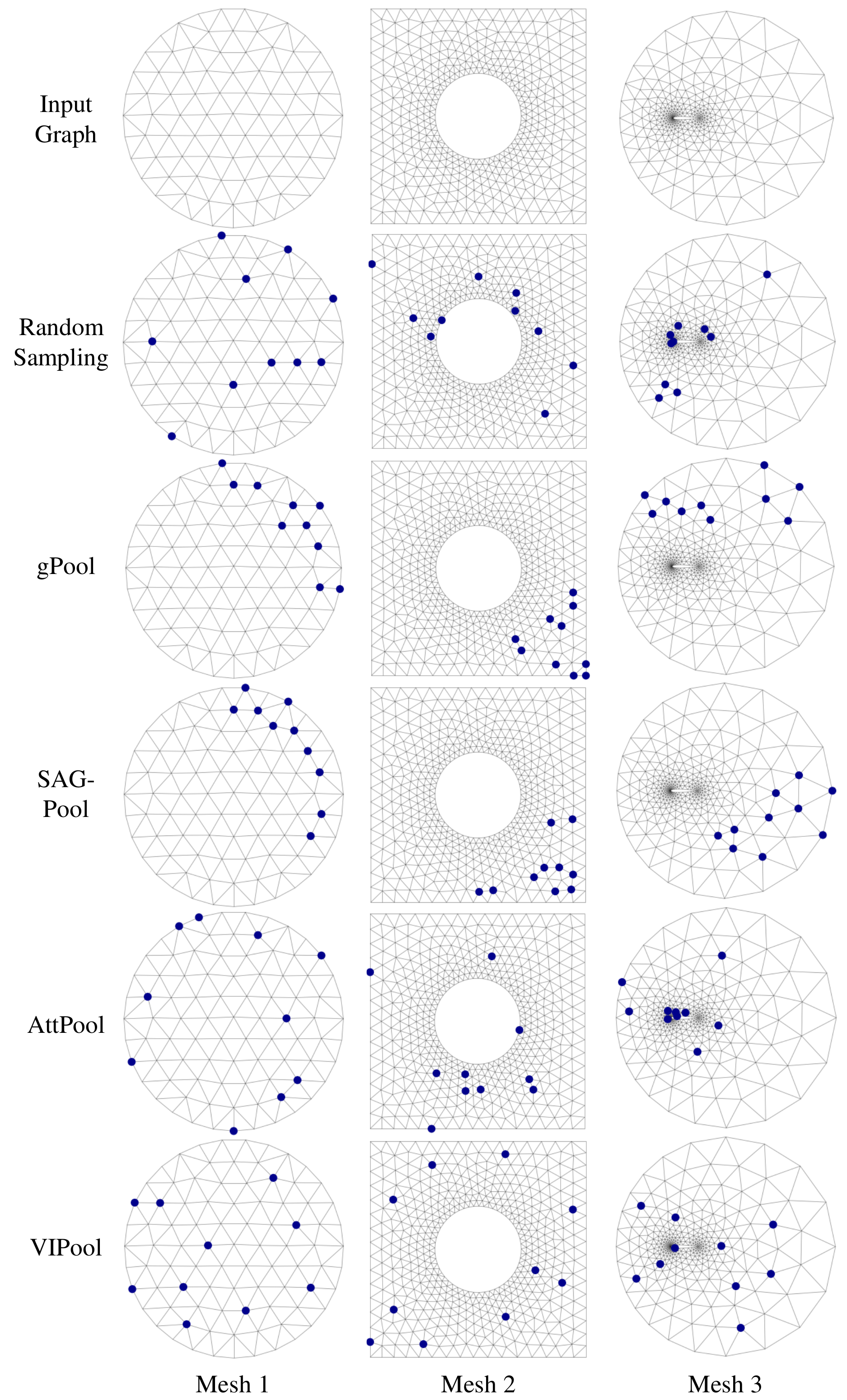}
    \caption{Vertex selection by using different pooling algorithms on three spatial mesh graph (better viewed on a color screen).}
    \label{fig:meshpool}
\end{figure}

The three mesh graphs have vertex features as the 2D Euclidean coordinates and the specific vertex distributions are that 1) 88 vertices uniformly distribute in a circle region; 2) 503 vertices distribute in a hollow square region, where the vertices densely distribute around the center and sparsely distribute near the margins; 3) 310 vertices distribute in a circle, where the vertices densely distribute near the center and sparsely distribute around. The specific topologies are shown in the first row of Figure~\ref{fig:meshpool}. 

We compare the proposed VIPool operation with several baseline methods: random sampling, gPool~\cite{ICML2019_Gao}, SAGPool~\cite{ICML2019_Lee} and AttPool~\cite{Huang_2019_ICCV}. The selected vertices are colored blue and illustrated in Figure~\ref{fig:meshpool}.
We see that, VIPool can abstract the original graphs more properly, where the preserved vertices distribute dispersely in both dense and sparse regions to cover the overall graphs. As for the baselines, we see that, 1) random sampling tends to select more vertices in dense regions, since each vertex is sampled with equal probability and the dense regions include more vertices and chances for vertex selection; 2) gPool and SAGpool calculate the importance weight for each vertices mainly based on vertex information itself without topological constraints, thus the selected vertices tends to distributed concentrated in local regions. 3) AttPool considers to model the local attentions and select more representative vertices, thus it can abstract graph structures to some extent, but the vertex distributions still slightly collapse the dense region.

% \newpage
% {\small
% \bibliographystyle{plainnat}
% \bibliography{egbib}
% }

% \end{document}

\end{document}